\documentclass{article} 
\usepackage{iclr2027_conference,times}

\iclrfinalcopy
\renewcommand{\headrulewidth}{0pt}

\usepackage{amsmath,amsfonts,bm}

\def\eqref#1{equation~\ref{#1}}

\def\1{\bm{1}}

\DeclareMathAlphabet{\mathsfit}{\encodingdefault}{\sfdefault}{m}{sl}
\SetMathAlphabet{\mathsfit}{bold}{\encodingdefault}{\sfdefault}{bx}{n}

\usepackage{hyperref}
\usepackage{url}

\usepackage{amsthm}
\usepackage{booktabs}
\usepackage{algorithm}
\usepackage{algpseudocode}
\usepackage{arydshln}
\usepackage{xcolor}
\definecolor{gainGreen}{RGB}{34,139,94}
\newcommand{\revision}[1]{#1}
\newtheorem{theorem}{Theorem}
\newtheorem{corollary}[theorem]{Corollary}
\usepackage{graphicx}

\title{PRICE the Action Chunks: Physical Relational Credit Assignment for Embodied Reinforcement Learning}

\author{%
\begin{minipage}[t]{\dimexpr\textwidth-2\tabcolsep\relax}
\centering
\fontsize{11}{13}\selectfont\normalfont
\mbox{Yangang Zou\textsuperscript{1,2,*}}, \mbox{Jiajun Lu\textsuperscript{3,*}},
\mbox{Weitao Zhou\textsuperscript{4,6,$\dagger$,$\ddagger$}}, \mbox{Haibao Yu\textsuperscript{5,7,$\ddagger$}},
\mbox{Bozhou Zhang\textsuperscript{1}},\\
\mbox{Jiawei Wang\textsuperscript{4}},
\mbox{Honglong Tian\textsuperscript{4}}, \mbox{Minglei Li\textsuperscript{4}},
\mbox{Li Zhang\textsuperscript{1,2,$\ddagger$}}\\[0.5em]
\normalfont\small
\textsuperscript{1}School of Data Science, Fudan University\quad
\textsuperscript{2}Shanghai Innovation Institute\quad
\textsuperscript{3}Beihang University\\
\textsuperscript{4}Simple AI\quad
\textsuperscript{5}Tuojing Intelligence\quad
\textsuperscript{6}Tsinghua University\quad
\textsuperscript{7}The University of Hong Kong\\[0.5em]
\footnotesize
\textsuperscript{*}Equal contribution.\quad
\textsuperscript{$\dagger$}Project leader.\quad
\textsuperscript{$\ddagger$}Corresponding authors.
\end{minipage}%
}

\begin{document}

\maketitle
\fancyhead{}

\begin{abstract}
Outcome-based reinforcement learning (RL) post-trains vision--language--action
policies using terminal success signals, but assigns the same trajectory-level
advantage to every action chunk. A failed episode can thus penalize useful early actions as if they caused the failure.
Existing approaches seek finer-grained feedback through learned evaluators, adding task-specific supervision or additional model
training. We explore, for the first time to our knowledge, whether physical
relations across trajectories can provide action-chunk credit in embodied
RL from terminal outcomes alone, without an auxiliary evaluator.
The key insight is that rollouts reaching corresponding physical situations
can serve as references for one another: their terminal outcomes provide
evidence for assessing local progress. We introduce \emph{Physical Relations for Inferring Credit
from Episodes} (PRICE), with two components:
(i) a physical relational graph that pools current and historical outcomes
at corresponding chunk boundaries to estimate success potentials; and
(ii) confidence-gated credit assignment that uses changes in these potentials
to refine trajectory-level supervision. Our analysis connects oracle
potential changes to the terminal-success objective and \revision{provides a
finite-sample directional bound for outcome-independent evidence pools}.
Independent continuation tests show that PRICE's retained credits align with local progress, while experiments on LIBERO, RoboTwin~2.0, and real robots demonstrate improved task success over outcome-based baselines and faster learning.
\end{abstract}

\section{Introduction}
\label{sec:introduction}

Vision--language--action (VLA) models learn general-purpose manipulation
policies from large-scale imitation data. Reinforcement learning (RL)
provides a way to refine these initial capabilities through interaction
\citep{kim2024openvla,tan2025interactive,li2026simplevla}.
Outcome-based approaches make this post-training practical by using task
success as the reward: a completed rollout supplies the feedback needed
for learning, without requiring annotations or reward design for each
intermediate manipulation stage.
 
Long-horizon manipulation, however, contains action chunks with different
effects on the final outcome. A robot may grasp an object securely and
transport it to the target, then fail only during placement. The failed
episode contains useful behavior as well as a decision that needs correction,
but a trajectory-level objective assigns the same advantage to every chunk.
Its terminal label tells the learner that the attempt failed, while leaving
unresolved where progress was made or lost. This makes local credit
assignment a central challenge for learning from sparse outcomes.

\begin{figure}[!t]
    \centering
    \includegraphics[width=\linewidth]{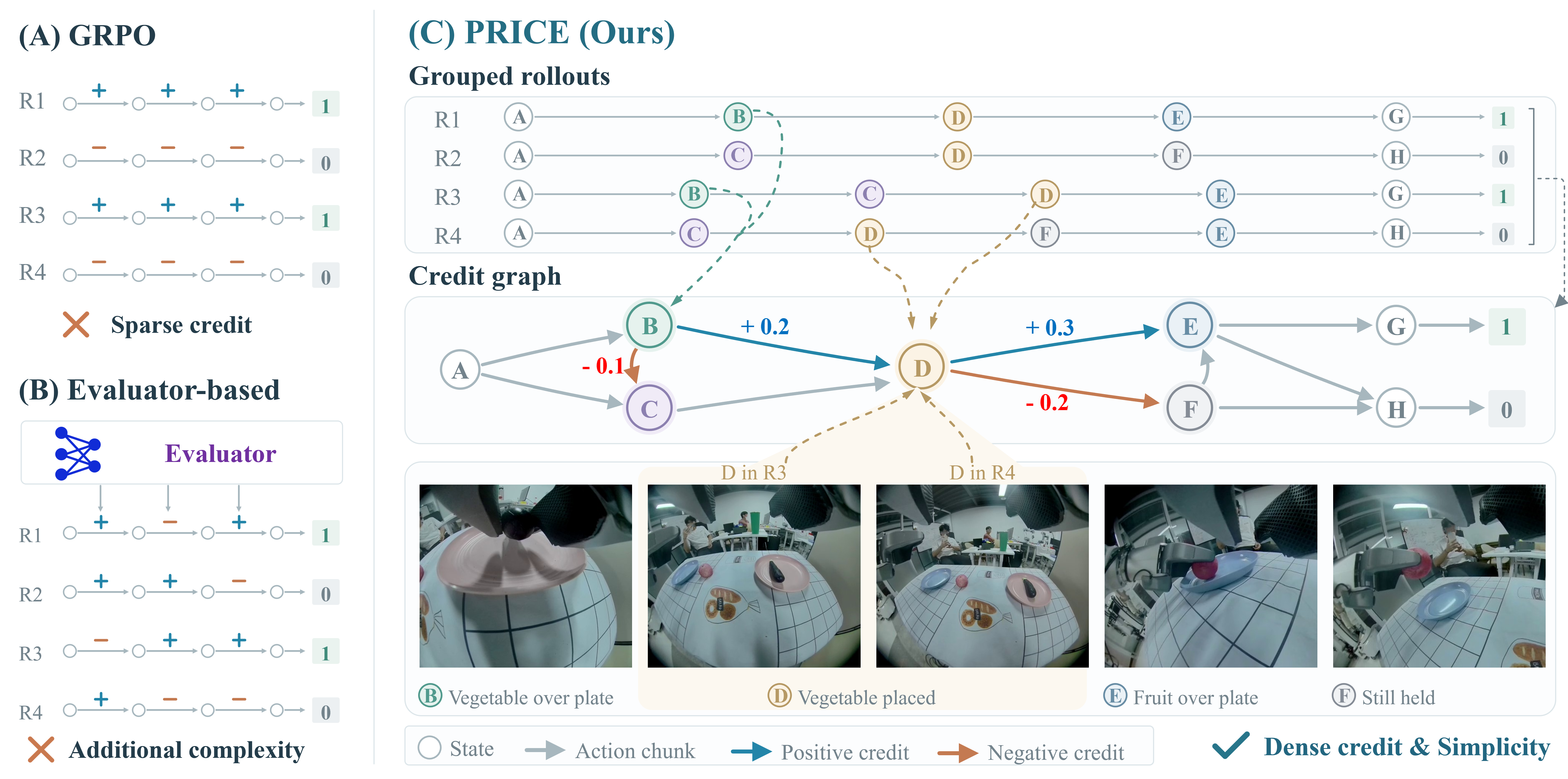}
    \vspace{-4pt} 
    \caption{Motivation for PRICE. PRICE derives local action-chunk credit by
    comparing physically corresponding states across rollouts, using only
    terminal outcomes and no auxiliary evaluator.}
    \label{fig:price-motivation}
    \vspace{-2pt}
\end{figure}

Existing approaches address this missing feedback through designed process
rewards, learned reward or value models, and world-model-based progress
estimation \citep{lu2025vla,shu2025rftf,xiao2025world,zhu2026wmpo}.
These mechanisms supply finer-grained supervision, but require task-specific
reward design or additional learned evaluators. This motivates our question:
\emph{Can embodied RL assign fine-grained credit to action chunks from terminal outcomes alone, without introducing an additional evaluator?}

To investigate this question, we revisit the experience generated by
group-based RL. Rollouts for the same task often pass through corresponding
physical situations before diverging in their later actions and outcomes.
For example, different attempts may establish a similar grasp, yet differ
in whether the object remains stable during transport. Such shared
situations allow trajectories to serve as references for one another.
Their terminal outcomes provide evidence about which intermediate
situations are more conducive to eventual success, making local progress
visible even within a failed episode. Figure~\ref{fig:price-motivation}
illustrates how shared and diverging manipulation attempts connect sparse
outcomes to local decisions.

The key challenge is to establish which physical situations can be compared
across trajectories. Cross-trajectory credit assignment has been explored in
agentic RL \citep{feng2026group,wang2026group,wang2026bipace,yan2026gated};
in embodied control, however, the relevant correspondence must be inferred
from continuous visual and proprioceptive observations. States that look
similar can differ in object pose or contact, with different implications
for subsequent actions. Useful relations must therefore capture physical
conditions that matter for task progress.

We therefore introduce \emph{Physical Relations for Inferring Credit from
Episodes} (PRICE) to derive action-chunk credit from physically grounded
cross-trajectory relations. PRICE identifies corresponding physical
situations from visual and proprioceptive observations and estimates their
success potentials using outcomes from current and historical rollouts.
Changes in these potentials across an action chunk provide local credit,
which is selected through confidence gating and added to the GRPO advantage.
This supervision comes directly from collected experience, without an auxiliary evaluator.
Our analysis connects
oracle potential changes to the terminal-success objective and \revision{provides
a finite-sample directional bound for outcome-independent evidence pools}.
To our knowledge, this is the first study to investigate action-chunk-level
credit assignment from physically grounded cross-trajectory relations in
embodied RL.

Empirically, PRICE narrows the performance gap between evaluator-free and
evaluator-based RL. It outperforms evaluator-free baselines  and learns faster than SimpleVLA-RL, while remaining competitive
with SRPO, a state-of-the-art evaluator-based method \citep{fei2026srpo}.
Independent continuation evaluation shows that PRICE credit aligns with local
progress measured by changes in node-averaged success. Controlled
ablations establish the contributions of historical evidence
and confidence gating to policy improvement. These benefits extend to
real-robot manipulation, where PRICE-AWR improves offline flow-matching
adaptation over trajectory-level GRPO-AWR.

Our contributions are threefold:
(i) We introduce PRICE, a method for inferring action-chunk credit from
physical relations and terminal outcomes without auxiliary evaluators.
(ii) We establish the connection between oracle endpoint credit and the
terminal-success objective, and analyze the directional reliability of
pooled credit estimates under explicit evidence-selection assumptions.
(iii) We demonstrate improved simulated and real-robot manipulation, with
independent credit evaluation and controlled component ablations.

\section{Related Work}
\label{sec:related-work}

\subsection{RL Post-Training for Embodied Policies}

Vision--language--action policies acquire broad manipulation capabilities from
imitation data, and reinforcement learning can further correct imitation errors
through interaction \citep{kim2024openvla}. Recent work extends post-training
across policy architectures and optimization frameworks: RIPT-VLA constructs
leave-one-out advantages directly from binary success, SimpleVLA-RL adopts
outcome-based GRPO, RLinf-VLA provides unified infrastructure for embodied RL,
and WAM-RL extends online adaptation to world--action models
\citep{tan2025interactive,li2026simplevla,zang2025rlinf,qian2026wam}.
Together, these studies establish the value of interactive post-training.
Outside VLA post-training, Q-Chunking treats an action chunk as the decision
unit and uses multi-step temporal-difference backups for sparse-reward
manipulation \citep{li2026reinforcement}.
Outcome-based variants, however, typically assign the same episode result to
every decision in a trajectory, revealing which rollout failed but not where
its probability of success changed.

\subsection{Fine-Grained Feedback in Embodied Reinforcement Learning}

To obtain finer feedback, one family introduces task-dependent process
supervision: TGRPO constructs dense rewards from intermediate task conditions,
VLA-RL learns a process reward model from automatically extracted manipulation
stages, and RFTF learns value changes by temporally ordering successful
demonstrations \citep{chen2025tgrpofinetuningvisionlanguageactionmodel,lu2025vla,shu2025rftf}. Feat2Go derives
visual progress targets from a pretrained world model and trains a value model
to reshape terminal rewards; PACE uses a phase-aware critic to assign step-level
credit; and Dream2Reward learns a successful-transition model from positive
demonstrations to score observed motion
\citep{shu2026feat2govisualfeaturegroundedvalue,song2026pace,zhang2026dream2reward}. Other world-model
approaches generate intermediate supervision through imagined future
interaction: World-Env combines an action-conditioned visual simulator with a
vision--language reflector for continuous reward and termination prediction;
WMPO couples pixel-space imagined rollouts with a learned reward model for
on-policy optimization; RISE decomposes its world model into controllable
dynamics and a progress-value model, scoring imagined actions by predicted
value improvement; and WoVR uses keyframe-started rollouts and
policy--simulator co-evolution to limit hallucination and accumulated model
error
\citep{xiao2025world,zhu2026wmpo,yang2026rise,jiang2026wovr}. SRPO uses the
policy's own successful rollouts as references and scores failures by their
distance to the success set, avoiding expert reference trajectories
\citep{fei2026srpo}. Temporal GRPO instead compares rollouts within detectable
task stages and assigns stage-relative advantages to their action intervals;
its stage construction and alignment use task-stage predicates during training
\citep{zhou2026temporal}. These approaches rely on task-specific progress
conditions or learned evaluators, while SRPO assigns progress at the rollout
level. This leaves open whether final task outcomes from grouped rollouts can
provide action-chunk credit without task-stage predicates or an auxiliary
evaluator.

\subsection{Credit Assignment from Cross-Trajectory Relations}

Credit assignment from delayed outcomes has broader precedents. RUDDER
redistributes delayed returns through a learned return decomposition
\citep{arjona2019rudder}. One-step value changes also underlie generalized
advantage estimation and potential-based reward shaping
\citep{schulman2015high,ng1999policy}. Recent agentic RL methods
propagate outcome signals through
repeated states, shared edges, or non-parametric node values, and by using
latent-space clustering to estimate local $Q-V$ signals from neighboring
rollouts
\citep{li2026salt,feng2026group,zhu2026gagpo,cheng2026beyond,wang2026group,yan2026gated,wang2026bipace}.
These studies show that cross-trajectory relations can provide local credit
without predefined progress labels. PRICE estimates value changes from pooled
terminal outcomes, and the policy-invariance result for potential-based reward
shaping does not directly apply to its estimated, gated score weights.
Relational credit, however, presupposes
knowing when different trajectories have reached the ``same state.'' In text
or reasoning tasks, this can often be approximated using discrete states,
repeated anchors, or semantic matching. In embodied control, two semantically
similar frames can differ in gripper--object contact or relative pose, while
the same physical situation can appear different under viewpoint or appearance
changes. Incorrectly merging or splitting such states directly distorts node
values and their differences
\citep{abel2016near,gelada2019deepmdp}. The central problem for embodied
relational credit is therefore to establish reliable cross-trajectory physical
correspondence from continuous visual and proprioceptive signals, and thereby
determine whether an action chunk makes genuine progress. PRICE constructs
physical relations at chunk boundaries, estimates local progress by comparing
the terminal-success potentials of preceding and succeeding endpoints across
trajectories, and \revision{selects credits using endpoint differences and
evidence counts} with a Hoeffding-style uncertainty threshold
\citep{hoeffding1963probability}.

\section{Preliminaries}
\label{sec:preliminaries}

\paragraph{Problem setup.}
For task context $\kappa$, let $H_t$ denote the complete interaction history
at action-chunk boundary $t$, and let $h_t$ denote the policy input
constructed from this history. The policy samples an action chunk
$\mathbf a_t\sim\pi_\theta(\cdot\mid h_t)$, whose execution produces
$H_{t+1}$. A trajectory
$\tau=(H_1,\mathbf a_1,\ldots,\mathbf a_T,H_{T+1})$ receives only a terminal
success indicator $Y\in\{0,1\}$. The post-training objective is
$J(\theta)=\mathbb E_{\kappa,\tau\sim\pi_\theta}[Y]$.
Let $T_i\leq T$ be the number of chunks executed in trajectory $i$.
  For analysis, early termination is padded with absorbing histories and
  policy-independent dummy actions.

\paragraph{Outcome-based GRPO.}
  Outcome-based GRPO \citep{shao2024deepseekmath} samples a group $g$ of $K$
  trajectories from the same initial context under a frozen collection policy
  $\pi_{\theta_{\mathrm{old}}}$. It assigns trajectory $i$ the
  group-normalized terminal advantage
  \begin{equation}
      \widehat A_i^{\mathrm{GRPO}}
      =\frac{Y_i-\bar Y_g}{s_{Y,g}+\varepsilon_{\mathrm{num}}},
      \qquad i\in g,
      \label{eq:grpo-trajectory-advantage}
  \end{equation}
  where $\bar Y_g$ and $s_{Y,g}$ are the group mean and sample standard
  deviation, and $\varepsilon_{\mathrm{num}}>0$ stabilizes normalization.
  If all outcomes in a group agree, every advantage is zero.

  For tokenized action chunks, GRPO maximizes the clipped surrogate
  \begin{equation}
      \begin{aligned}
      \mathcal J_{\mathrm{GRPO}}(\theta)
      =\mathbb E\Bigg[
      &\frac{1}{N_{\mathrm{tok}}}
      \sum_{i,t,\ell}
      \min\!\Big(
        \rho_{i,t,\ell}(\theta)\widehat A_i^{\mathrm{GRPO}},\\
      &\operatorname{clip}\!\left(
        \rho_{i,t,\ell}(\theta),1-\epsilon_{\mathrm{low}},
        1+\epsilon_{\mathrm{high}}\right)
        \widehat A_i^{\mathrm{GRPO}}
      \Big)
      -\beta D_{\mathrm{KL}}(\pi_\theta\|\pi_{\mathrm{ref}})
      \Bigg].
      \end{aligned}
      \label{eq:grpo-surrogate}
  \end{equation}
  Here $\rho_{i,t,\ell}$ is the current-to-collection policy likelihood
  ratio for a valid action token, and
  $N_{\mathrm{tok}}=\sum_{i,t}L_{i,t}$ counts valid tokens, with $L_{i,t}$
  the number in chunk $t$ of trajectory $i$. The clipping bounds are
  $\epsilon_{\mathrm{low}}$ and $\epsilon_{\mathrm{high}}$; $\beta$ weights
  an optional reference-policy KL penalty. Every chunk in a trajectory
  receives the same advantage, leaving its local contribution unresolved.

\section{Method}
\label{sec:method}

\begin{figure}[t]
    \centering
    \includegraphics[width=\linewidth]{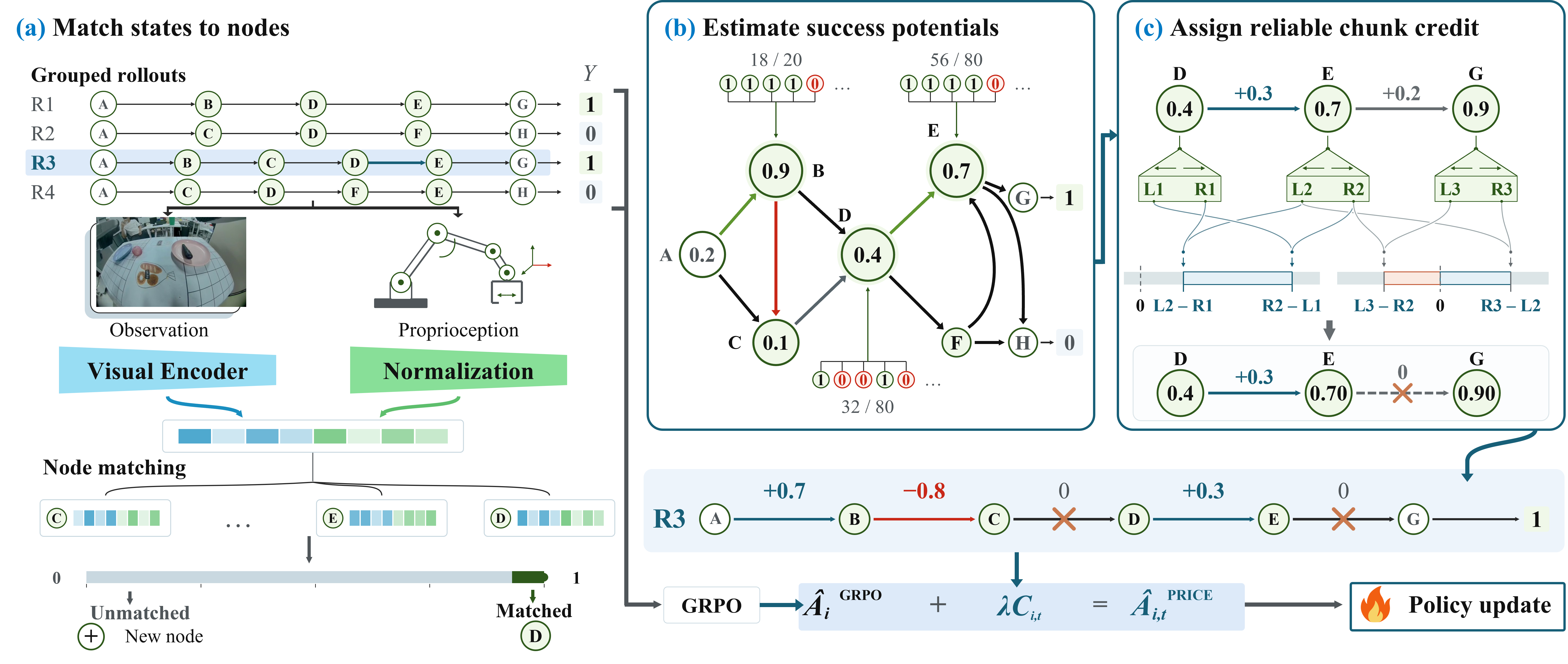}
    \caption{Overview of PRICE.
    (a) Visual--proprioceptive descriptors match physically corresponding
    states across rollouts.
    (b) Cross-trajectory terminal outcomes determine node success potentials.
    (c) Confidence-gated endpoint differences provide action-chunk credit
    that augments the GRPO advantage for policy optimization.}
    \label{fig:price-method}
\end{figure}

Given grouped rollouts labeled only by terminal success, PRICE maps action-chunk
boundaries to task-specific physical nodes, estimates endpoint success
potentials from cross-trajectory evidence, and converts changes in graph
potential into confidence-gated action-chunk credit that augments the
trajectory-level GRPO advantage, as shown in Figure~\ref{fig:price-method}.

We seek to credit action chunks that increase the probability of eventual
success. This motivates measuring local progress by the change in success
probability across chunk boundaries. For a fixed policy $\pi$, let
$V_t^\pi(H_t)=\mathbb E_\pi[Y\mid H_t]$ denote this
probability given history $H_t$, and define
the chunk advantage
$A_t^\pi(H_t,\mathbf a_t)=\mathbb E_\pi[Y\mid H_t,\mathbf a_t]-V_t^\pi(H_t)$.
We define the oracle chunk credit as
$\Delta_t^{\mathrm{ora}}=V_{t+1}^\pi(H_{t+1})-V_t^\pi(H_t)$, with
$V_{T+1}^\pi(H_{T+1})=Y$.

\begin{theorem}[Oracle alignment of additive credit]
\label{thm:oracle-objective-alignment}
Assume a finite-horizon process whose task and initial-context distribution,
environment transitions, and terminal-outcome law do not depend on $\theta$,
together with the standard regularity conditions for likelihood-ratio policy
gradients. For $\pi=\pi_\theta$,
\begin{equation}
    \mathbb E_\pi[\Delta_t^{\mathrm{ora}}\mid H_t,\mathbf a_t]
    =A_t^\pi(H_t,\mathbf a_t).
    \label{eq:oracle-progress-advantage}
\end{equation}
For on-policy grouped rollouts, let
$S_t=\nabla_\theta\log\pi_\theta(\mathbf a_t\mid h_t)$ and
$g_{\mathrm{base}}=\mathbb E_\pi[\sum_t S_t\widehat A^{\mathrm{GRPO}}]$.
With stop-gradient weights and $\lambda\geq0$, the oracle score update satisfies
\begin{equation}
    g_{\mathrm{aug}}^{\mathrm{ora}}
    :=\mathbb E_\pi\!\left[\sum_t S_t
        \bigl(\widehat A^{\mathrm{GRPO}}+\lambda\Delta_t^{\mathrm{ora}}\bigr)
      \right]
    =g_{\mathrm{base}}+\lambda\nabla_\theta J(\theta).
    \label{eq:oracle-additive-alignment}
\end{equation}
\end{theorem}
The theorem shows that weighting policy scores by oracle potential
differences recovers the gradient of the terminal-success objective in
expectation. Motivated by this connection, PRICE constructs local credit
from differences in pooled success potentials at physically matched
boundaries and uses confidence gating to \revision{select differences relative
to their estimated uncertainty}.
Appendix~\ref{app:advantage-identity} gives the proof.

\subsection{Physical Relational Graph Construction}
\label{sec:physical-graph}

\paragraph{Physical Interaction Representation.}
Let $i$ index a trajectory and $t$ an action-chunk boundary. PRICE combines the
visual observation $o_{i,t}$ and robot proprioception $p_{i,t}$ at that boundary
into a physical interaction descriptor
\begin{equation}
    z_{i,t}
    =\left[
      \sqrt{w_{\mathrm{vis}}}\,
      \mathcal N\!\left(\phi_{\mathrm{vis}}(o_{i,t})\right);
      \sqrt{w_{\mathrm{prop}}}\,
      \mathcal N\!\left(\phi_{\mathrm{prop}}(p_{i,t})\right)
    \right],
    \quad
    w_{\mathrm{vis}},w_{\mathrm{prop}}\geq0,
    \quad
    w_{\mathrm{vis}}+w_{\mathrm{prop}}=1,
    \label{eq:physical-representation}
\end{equation}
where $\mathcal N$ denotes $\ell_2$ normalization,
$\phi_{\mathrm{vis}}$ is a frozen visual encoder, and
$\phi_{\mathrm{prop}}$ applies fixed normalization to proprioception.
This fixed representation keeps node matching comparable across RL updates.
The resulting nodes form an observation-based abstraction of interaction
histories, over which continuation evidence is pooled.

\paragraph{Node matching and relation edges.}
For each task $\kappa$, PRICE maintains a directed graph shared across its
initial-condition cases. Nodes cluster similar boundary descriptors, and
edges represent action chunks.
The prototype $c_v$ of node $v$ is the normalized sum of its assigned
descriptors. In round $q$, PRICE compares $z_{i,t}$
with every node prototype frozen at the start of the round using cosine
similarity $\langle z_{i,t},c_v\rangle$. If the highest
similarity reaches the matching threshold $\eta$, the boundary is assigned to
the corresponding node. Unmatched descriptors are jointly clustered within
each rollout group using the same threshold $\eta$, forming
temporary nodes for that group's queries. Let $v_{i,t}$ denote a boundary's
node assignment. Once two consecutive boundaries are assigned,
$\mathbf a_{i,t}$ defines the directed relation
$(v_{i,t},\mathbf a_{i,t},v_{i,t+1})$.

\paragraph{Cross-trajectory node potentials.}
PRICE estimates each node's success potential by pooling terminal outcomes
across the histories represented by the node. Let $g(i)$ denote the rollout
group containing query trajectory $i$, whose trajectories share the same
initial context. Let
$I_j(v)=\mathbb{I}[\exists\,t:\ v_{j,t}=v]$ indicate whether trajectory $j$
visits node $v$. Each trajectory
contributes once per node, using its first visit as the representative history.
For query trajectory $i$ of task $\kappa(i)$, the current counts at node $v$ are
\begin{equation}
    \begin{aligned}
    n^{\mathrm{cur}}_{-i}(v)
    &=\sum_{j\in g(i),\,j\neq i}I_j(v),\\
    r^{\mathrm{cur}}_{-i}(v)
    &=\sum_{j\in g(i),\,j\neq i}I_j(v)Y_j.
    \end{aligned}
    \label{eq:cross-trajectory-node-value}
\end{equation}
These count contributing trajectories and successful outcomes, respectively.
When $n^{\mathrm{cur}}_{-i}(v)>0$, the current estimate is
$r^{\mathrm{cur}}_{-i}(v)/n^{\mathrm{cur}}_{-i}(v)$. Only trajectories matched
to the same node share evidence, and the query trajectory is excluded.
GRPO normalization remains within each rollout group.

\paragraph{Cross-round node potentials.}
Current-group evidence may be sparse or have little outcome variation.
PRICE therefore augments node estimates with historical
evidence. For each node, a summary $s$ stores per-round counts $(n_s,r_s)$ of
distinct visiting trajectories and successful outcomes, aggregated across
groups of the same task after each update. To limit mismatch from
policy changes, it uses only summaries satisfying $D_{s,q}\leq d_{\max}$, where $D_{s,q}$ is the
cumulative adjacent-policy KL from the summary's collection round $u_s$ to the
current round $q$, and $d_{\max}$ is the maximum allowed cumulative KL for
reusing historical evidence.

PRICE combines current-group and eligible historical counts for the same node.
For a boundary $x$ matched to node $v$, the pooled success potential is
\begin{equation}
    \widehat V_x
    =\frac{r^{\mathrm{cur}}_{-i}(v)+\sum_s r_s}
          {n^{\mathrm{cur}}_{-i}(v)+\sum_s n_s},
    \label{eq:bounded-history-node-value}
\end{equation}
where the sums run over eligible historical summaries for $v$. The denominator,
denoted by $N_x$, counts all contributing trajectories, giving each equal
weight. No estimate is formed when $N_x=0$.

\subsection{Confidence-Gated Action-Chunk Credit}
\label{sec:sampling-gate}

Finite-sample noise can reverse the sign of an estimated potential change.
PRICE \revision{uses a Hoeffding-inspired gate to compare this change with a
threshold determined by endpoint support}. For chunk $t$ in trajectory $i$, let $x$ and
$x^+$ denote its source and destination boundaries. When both estimates are
available, the candidate credit is
$\widehat\Delta_{i,t}=\widehat V_{x^+}-\widehat V_x$.

An endpoint is supported when it is assigned to a node and $N_x>0$. For a
chosen \revision{gate parameter} $\delta_{\mathrm{edge}}\in(0,1)$, its
Hoeffding radius and interval are
\begin{equation}
    \epsilon_x=
    \sqrt{\frac{\log(4/\delta_{\mathrm{edge}})}{2N_x}},
    \qquad
    L_x=\max(0,\widehat V_x-\epsilon_x),
    \qquad
    R_x=\min(1,\widehat V_x+\epsilon_x).
    \label{eq:pooled-hoeffding-interval}
\end{equation}
Appendix~\ref{app:pooled-endpoint-concentration} provides the derivation and
sampling assumptions.

Propagating the two endpoint intervals gives an interval for the potential
change:
\begin{equation}
    \mathcal I_{i,t}
    =\left[L_{x^+}-R_x,\;R_{x^+}-L_x\right].
    \label{eq:edge-sampling-interval}
\end{equation}

PRICE accepts the candidate credit only if this interval excludes zero:
\begin{equation}
    C_{i,t}=
    \begin{cases}
      \widehat\Delta_{i,t},
      &\text{if }t<T_i,\ x,x^+\text{ are supported, and }
       0\notin\mathcal I_{i,t},\\
      0,&\text{otherwise}.
    \end{cases}
    \label{eq:price-credit}
\end{equation}
Here $T_i$ is the number of chunks in trajectory $i$; its final chunk retains
the base GRPO advantage.

Let $\overline V_x$ denote the mean conditional continuation-success probability of the
pooled records under their respective collection policies, and define
$\overline\Delta_{i,t}=\overline V_{x^+}-\overline V_x$.

{
\paragraph{Gate interpretation.}
For edge $e=(i,t)$, write $r_e=\epsilon_x+\epsilon_{x^+}$ and
$E_e=|\widehat V_x-\overline V_x|
     +|\widehat V_{x^+}-\overline V_{x^+}|$.
An accepted credit satisfies $|\widehat\Delta_e|>r_e$. Since
$|\widehat\Delta_e-\overline\Delta_e|\leq E_e$, the triangle inequality gives
\begin{equation}
    C_e\neq0,\quad E_e\leq r_e
    \quad\Longrightarrow\quad
    \operatorname{sign}(C_e)=\operatorname{sign}(\overline\Delta_e).
    \label{eq:gate-error-condition}
\end{equation}
Thus, a wrong nonzero credit requires $E_e>r_e$, for any archive selection.
Appendix~\ref{app:sampling-gate-proof} gives a finite-sample bound under
outcome-independent evidence selection, and Appendix~\ref{app:credit-transfer}
relates these errors to collection-policy mismatch and state abstraction.
\par}

\subsection{Optimization Objective}
\label{sec:price-optimization}

PRICE augments the trajectory-level GRPO advantage with confidence-gated
action-chunk credit:
\begin{equation}
    \widehat A_{i,t}^{\mathrm{PRICE}}
    =\widehat A_i^{\mathrm{GRPO}}+\lambda C_{i,t},
    \qquad \lambda\geq0.
    \label{eq:price-augmented-advantage}
\end{equation}
Here $\lambda$ controls the strength of local credit. We optimize the clipped
objective in Equation~\ref{eq:grpo-surrogate}, assigning
$\widehat A_{i,t}^{\mathrm{PRICE}}$ to every valid action token in chunk $t$
of trajectory $i$. Implementation details are provided in
Appendix~\ref{app:training-lifecycle}.

\section{Experiments}
\label{sec:experiments}

\subsection{Experimental Setup}
\label{sec:experimental-setup}

\paragraph{Benchmarks.}
We evaluate PRICE on two simulated benchmarks and two real-robot tasks.
For LIBERO \citep{liu2023libero}, we use the Spatial, Object, Goal, and Long
suites. From RoboTwin~2.0 \citep{chen2025robotwin}, we select six
manipulation tasks on a dual-arm platform covering grasping,
     coordinated lifting, object transport, placement, and handover.
For evaluation, we follow the RoboTwin protocol used by RLinf-VLA
\citep{zang2025rlinf}.
Real-robot experiments use Remote Insertion and Produce Sorting from
HiFi-UMI \citep{ai2026hifi}, covering precise placement and
category-conditioned sorting (Section~\ref{sec:real-world-experiments}).

\paragraph{Implementation details.}
For LIBERO, we use OpenVLA-OFT initialized by one-shot SFT.
  For RoboTwin~2.0, we use OpenVLA-OFT with task-specific SFT
  checkpoints through RLinf~\citep{zang2025rlinf}. We fine-tune all parameters on eight NVIDIA A800
GPUs with 80\,GB memory each, using a learning rate of $5\times10^{-6}$
and an optimization mini-batch size of 128. Each raw rollout batch contains
64 initial-condition cases with eight rollouts per case (512 trajectories),
sampled at a temperature of $1.6$. For our main PRICE experiments on LIBERO and RoboTwin~2.0, we
  report mean success rates over three independent training runs.
  The corresponding standard deviations are reported in
  Appendix~\ref{app:main-results-std}. PRICE-specific settings are provided in
Appendix~\ref{app:hyperparameters}. Real-robot experiments use the
offline flow-matching adaptation described in
Section~\ref{sec:real-world-experiments}.

\paragraph{Metrics.}
Following the official evaluation protocols, we report task success rate
(\%) and measure training efficiency by the steps needed to reach target
success rates. We assess local credit using independent Monte Carlo (MC)
estimates of the change in node-averaged continuation success across each
chunk. Direction accuracy (DirAcc, \%) measures sign agreement between
credit and MC changes on non-tied comparisons. Spearman rank correlation
(RankCorr) measures whether signed credits and MC changes rank chunks
consistently. Aligned MC gap (AlignedGap, pp) measures progress in the
credit's direction by averaging the MC change multiplied by the credit
sign. Higher is better for all three. The MC protocol and metric
definitions are given in Appendix~\ref{app:credit-evaluation-protocol}.

\subsection{Main Results}
\label{sec:main-results}

\paragraph{Task performance.}
Tables~\ref{tab:main-results} and~\ref{tab:robotwin_main} summarize task
performance on LIBERO and RoboTwin~2.0. On LIBERO, PRICE achieves $99.0\%$
average success, outperforming all compared evaluator-free methods and
closely matching SRPO ($99.2\%$). On LIBERO-Long, PRICE reaches $97.6\%$,
exceeding the strongest evaluator-free baseline by $3.6$ percentage points.
Across the six RoboTwin~2.0 tasks reported in
  Table~\ref{tab:robotwin_main}, PRICE achieves an average success rate
  of $90.0\%$, compared with $31.9\%$ for the OpenVLA-OFT (SFT)
  initialization, an improvement of $58.1$ percentage points. This
  average is numerically higher than Feat2Go ($88.8\%$), RLinf-VLA
  ($84.5\%$), and SimpleVLA-RL ($71.3\%$) by $1.2$, $5.5$, and $18.7$
  percentage points, respectively.

\begin{table}[t]
    \caption{Performance comparison on the LIBERO benchmark. We report task
    success rates (\%). Rows below the dashed line are our results;
    other values follow the unified comparison of \citet{fei2026srpo}.
    T, W, P, and I denote
    third-person images, wrist images, proprioception, and language
    instructions, respectively.
    Full and One denote full-shot and one-trajectory-per-task SFT, respectively.}
    \label{tab:main-results}
    \centering
    \scriptsize
    \setlength{\tabcolsep}{4.5pt}
    \begin{tabular}{lcccccc}
        \toprule
        Method & Policy input & Spatial & Object & Goal & Long & Avg. \\
        \midrule
        \multicolumn{7}{l}{\textit{VLA baseline}} \\
        OpenVLA~\citep{kim2024openvla}
            & T+I & 84.7 & 88.4 & 79.2 & 53.7 & 76.5 \\
        OpenVLA$^{\ast}$-Full~\citep{fei2026srpo}
            & T+I & 91.6 & 95.3 & 90.6 & 86.5 & 91.0 \\
        \midrule
        \multicolumn{7}{l}{\textit{Evaluator-based RL}} \\
        TGRPO~\citep{chen2025tgrpofinetuningvisionlanguageactionmodel}
            & T+I & 90.4 & 92.2 & 81.0 & 59.2 & 80.7 \\
        GRAPE~\citep{zhang2024grape}
            & T+I & 88.5 & 92.1 & 83.1 & 57.2 & 80.2 \\
        VLA-RL~\citep{lu2025vla}
            & T+I & 90.2 & 91.8 & 82.2 & 59.8 & 81.0 \\
        World-Env~\citep{xiao2025world}
            & T+I & 87.6 & 86.6 & 86.4 & 57.8 & 79.6 \\
        SRPO~\citep{fei2026srpo}
            & T+I & 98.8 & 100.0 & 99.4 & 98.6 & 99.2 \\
        \midrule
        \multicolumn{7}{l}{\textit{Evaluator-free RL}} \\
        SimpleVLA-RL~\citep{li2026simplevla}
            & T+I & 98.2 & 98.7 & 98.8 & 91.7 & 96.9 \\
        RIPT-VLA~\citep{tan2025interactive}
            & T+W+P+I & 99.0 & 98.6 & 98.6 & 93.8 & 97.5 \\
        RLinf-VLA~\citep{zang2025rlinf}
            & T+W+P+I & 99.4 & 99.8 & 98.8 & 94.0 & 98.0 \\
        \noalign{\vskip\aboverulesep}
        \hdashline
        \noalign{\vskip\belowrulesep}
        OpenVLA-OFT-One
            & T+I & 63.6 & 54.9 & 59.6 & 17.3 & 48.9 \\
        \textbf{$+$ PRICE (Ours)}
            & T+I & 99.2 & 99.6 & 99.4 & 97.6 & 99.0 \\
        & & \textcolor{gainGreen}{$\uparrow$35.6}
            & \textcolor{gainGreen}{$\uparrow$44.7}
            & \textcolor{gainGreen}{$\uparrow$39.8}
            & \textcolor{gainGreen}{$\uparrow$80.3}
            & \textcolor{gainGreen}{$\uparrow$50.1} \\
        \bottomrule
    \end{tabular}
\end{table}

\begin{table}[t]
    \caption{Performance comparison on six selected RoboTwin~2.0
  manipulation tasks. We report task success rates (\%) and the
  average across the six tasks. The OpenVLA-OFT (SFT) row is the
  initialization used for PRICE; results for the other compared
  methods are from the cited works.}
    \label{tab:robotwin_main}
    \centering
    \scriptsize
    \setlength{\tabcolsep}{3.2pt}
    \renewcommand{\arraystretch}{1.08}
    \begin{tabular}{lccccccc}
        \toprule
        Method
        & \shortstack{Handover\\Block}
        & \shortstack{Lift\\Pot}
        & \shortstack{Move Can\\Pot}
        & \shortstack{Pick Dual\\Bottles}
        & \shortstack{Place Container\\Plate}
        & \shortstack{Place Empty\\Cup}
        & Avg. \\
        \midrule
        SimpleVLA-RL~\citep{li2026simplevla}
        & 57.8 & 64.1 & 61.2 & 68.3 & 82.1 & 94.2 & 71.3 \\
        RLinf-VLA~\citep{zang2025rlinf}
        & 70.31 & 70.31 & 83.59 & 92.96 & 95.31 & 94.53 & 84.5 \\
        Feat2Go~\citep{shu2026feat2govisualfeaturegroundedvalue}
        & 76.6 & 78.9 & \textbf{91.4} & 93.8 & 96.9 & 95.3 & 88.8 \\
        \noalign{\vskip\aboverulesep}
        \hdashline
        \noalign{\vskip\belowrulesep}
        OpenVLA-OFT (SFT)
        & 28.1 & 3.1 & 9.4 & 20.3 & 54.7 & 75.8 & 31.9 \\
        \textbf{$+$ PRICE (Ours)}
        & \textbf{79.7}
        & \textbf{80.5}
        & 90.6
        & \textbf{94.5}
        & \textbf{97.7}
        & \textbf{96.9}
        & \textbf{90.0} \\
        & \textcolor{gainGreen}{$\uparrow$51.6}
        & \textcolor{gainGreen}{$\uparrow$77.4}
        & \textcolor{gainGreen}{$\uparrow$81.2}
        & \textcolor{gainGreen}{$\uparrow$74.2}
        & \textcolor{gainGreen}{$\uparrow$43.0}
        & \textcolor{gainGreen}{$\uparrow$21.1}
        & \textcolor{gainGreen}{$\uparrow$58.1} \\
        \bottomrule
    \end{tabular}
\end{table}

\begin{figure}[t]
    \centering
    \includegraphics[width=\linewidth]{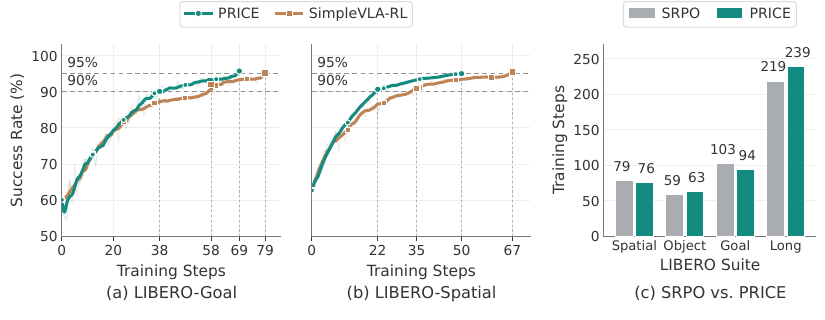}
    \caption{Training efficiency on LIBERO.
    (a--b) Rollout success during training on Goal and Spatial. Solid curves show
    seven-step moving averages and faint lines show raw observations;
    dashed guides and x-axis ticks mark the steps for 90\% and 95\% success.
    (c) Training steps of SRPO and PRICE across four suites.
    SRPO counts are from \citet{fei2026srpo};
    corresponding success rates are reported in Table~\ref{tab:main-results}.}
    \label{fig:training-efficiency}
\end{figure}

\paragraph{Training efficiency.}
\label{sec:training-efficiency}
We compare PRICE with SimpleVLA-RL under the same settings.
PRICE reaches both $90\%$ and $95\%$ success with fewer training steps and
lower GPU-hour costs than SimpleVLA-RL
(Figure~\ref{fig:training-efficiency}(a--b)). At $90\%$ success, PRICE reduces
training steps by $34.5\%$ on Goal and $37.1\%$ on Spatial, with corresponding
GPU-hour savings of $38.3\%$ and $32.8\%$. Timing details are provided in
Appendix~\ref{app:training-gpu-hours}. PRICE also achieves average
success comparable to evaluator-based SRPO under similar training budgets
(Table~\ref{tab:main-results}; Figure~\ref{fig:training-efficiency}(c)).

\subsection{Real-World Experiments}
\label{sec:real-world-experiments}

\begin{figure}[t]
    \centering
    \includegraphics[width=\linewidth]{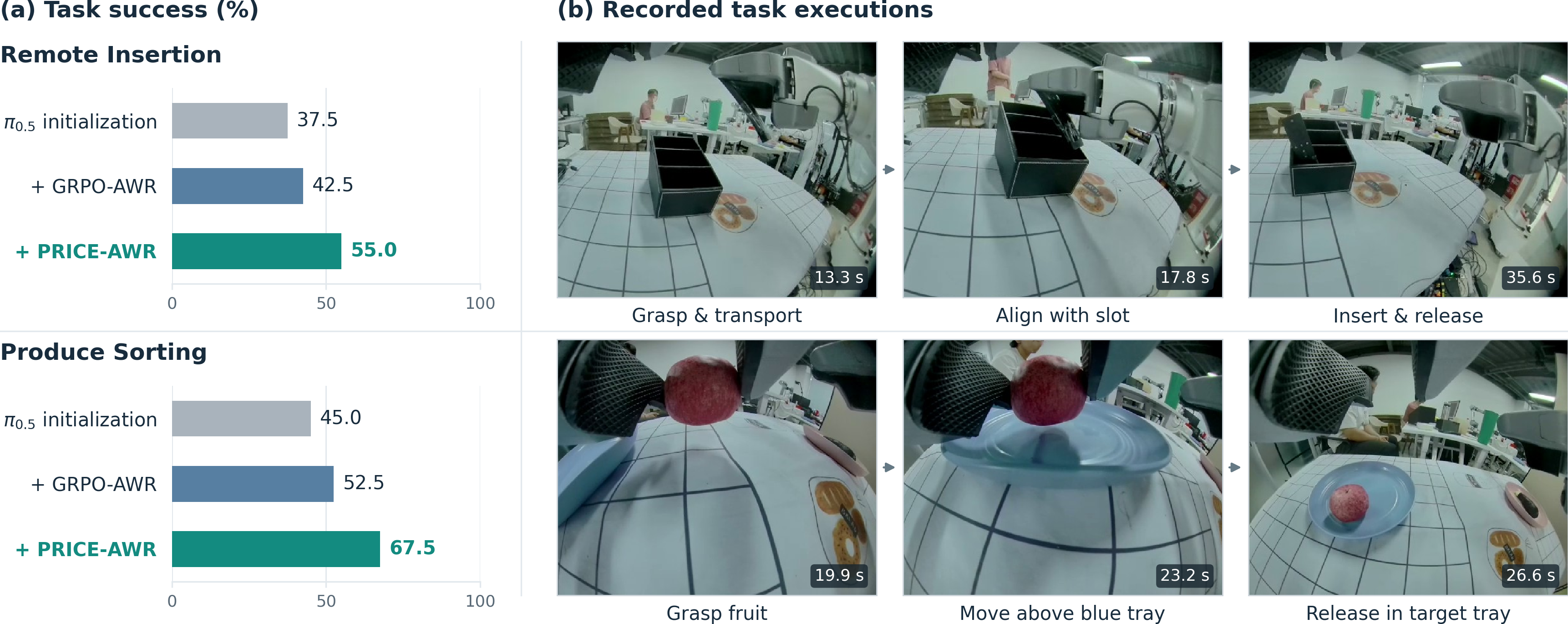}
    \caption{Real-robot performance and representative executions.
  (a) Success rates of the initialization, GRPO-AWR, and PRICE-AWR
  on Remote Insertion and Produce Sorting under the HiFi-UMI
  evaluation protocol.
  (b) Chronological keyframes show remote transport, alignment, and
  insertion into the storage box (top), and fruit grasping, transport,
  and release into its designated tray during Produce Sorting (bottom).}
    \label{fig:real-robot-composite}
\end{figure}

\paragraph{Experimental setup.}
We evaluate on two HiFi-UMI tasks \citep{ai2026hifi}: Remote Insertion
places a remote control into a storage box, while Produce Sorting requires
placing both a fruit and a vegetable into their category-specific trays
(Figure~\ref{fig:real-robot-composite}(b)). All variants start from the same
OpenPI-$\pi_{0.5}$ checkpoint post-trained on 400 HiFi-UMI trajectories.
The initialization collects 60 rollouts per task for offline adaptation,
with no additional online collection during optimization. We compare the
initialization with two offline adaptation methods based on
advantage-weighted regression (AWR)~\citep{peng2019advantage}.
GRPO-AWR weights the flow-matching loss
using trajectory-level advantages, while PRICE-AWR adds action-chunk
relational credit before computing these weights.
Both adaptation methods share the same data, random
seed, and optimization budget. We follow the HiFi-UMI evaluation protocol,
reporting task success and the macro-average across tasks.
Appendix~\ref{app:real-robot-details} details the adaptation objectives,
observation configuration, and training hyperparameters.

\paragraph{Results.}
PRICE-AWR achieves $55.0\%$ success on Remote Insertion and $67.5\%$ on
Produce Sorting, exceeding GRPO-AWR by $12.5$ and $15.0$ percentage points,
respectively (Figure~\ref{fig:real-robot-composite}(a)). Macro-average
success reaches $61.25\%$, compared with $47.5\%$ for GRPO-AWR and $41.25\%$
for the initialization. These gains on both tasks support the benefit of
action-chunk relational credit under the same offline adaptation budget.

\subsection{Analysis of PRICE}
\label{sec:graph-credit-analysis}

\begin{table}[!htbp]
    \caption{Component ablations on LIBERO-Spatial and LIBERO-Object.
    All variants use 63 training steps with the same initialization,
    interaction budget, and optimization settings within each suite. We report task
    success (\%) and the average across the two suites.}
    \label{tab:ablations}
    \centering
    \small
    \begin{tabular}{lccccc}
        \toprule
        Variant & History & Gate & Spatial & Object & Avg. \\
        \midrule
        GRPO & $\times$ & $\times$ & 93.2 & 94.6 & 93.9 \\
        PRICE w/o Historical Evidence & $\times$ & $\checkmark$ & 93.4 & 94.6 & 94.0 \\
        PRICE w/o Gate & $\checkmark$ & $\times$ & 95.8 & 97.9 & 96.9 \\
        \textbf{Full PRICE} & $\checkmark$ & $\checkmark$
            & \textbf{97.0} & \textbf{99.6} & \textbf{98.3} \\
        \bottomrule
    \end{tabular}
\end{table}

\begin{figure}[t]
    \centering
    \includegraphics[width=\linewidth]{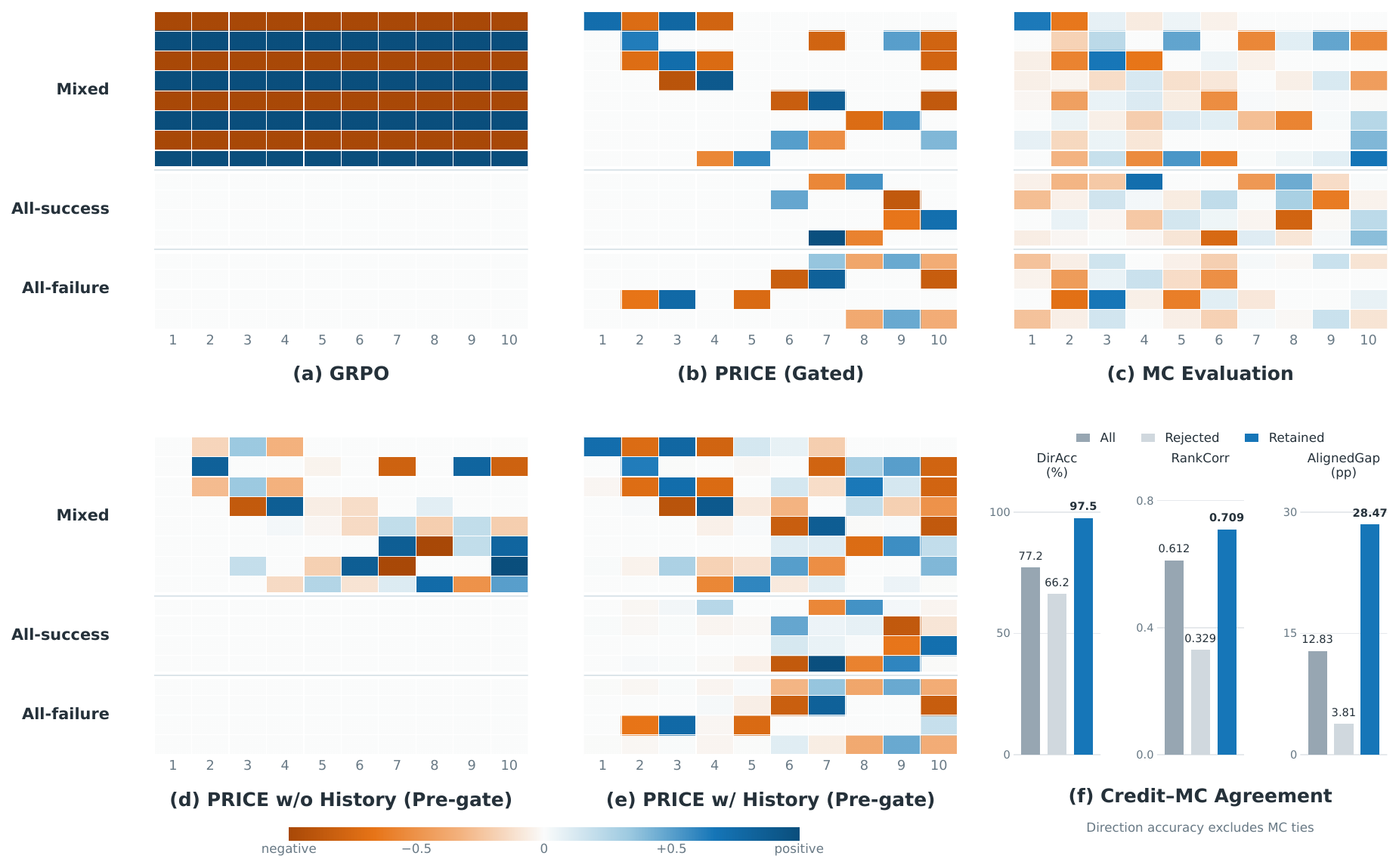}
    \caption{Credit analysis on LIBERO-Spatial.
    Rows denote trajectory windows and columns action chunks.
    (a) GRPO advantage sign. (b) Gated PRICE credit from (e).
    (c) Independent MC progress over the pooled evidence in (e).
    (d--e) Pre-gate contrasts without/with history at fixed nodes.
    (f) MC-based credit evaluation for all nonzero candidates in (e) and their
    rejected/retained subsets.}
    \label{fig:credit-analysis}
\end{figure}

\paragraph{Alignment with local progress.}
\label{sec:credit-alignment}
We independently assess local progress on LIBERO-Spatial by continuing the
  frozen OpenVLA-OFT policy from saved states in each chunk's source and
  destination evidence pools. The difference in node-averaged Monte Carlo
  success rates provides a reference for each chunk's progress
  (Appendix~\ref{app:credit-evaluation-protocol}). Figure~\ref{fig:credit-analysis}(a--c)
  shows that PRICE distinguishes chunks even when their trajectories share
  the same terminal outcome. Retained credits agree in sign with $97.5\%$
  of non-tied reference differences, with a RankCorr of $0.709$ and an
  AlignedGap of $+28.47$ pp (Figure~\ref{fig:credit-analysis}(f)).

\paragraph{Roles of historical evidence and confidence gating.}
\label{sec:credit-components}
Historical evidence enables local contrasts even when all trajectories in
a rollout group share the same outcome. Pooling outcomes across rounds
yields distinct endpoint potentials in these groups
(Figure~\ref{fig:credit-analysis}(d,e)). Confidence gating then selects a
subset with stronger agreement with independent MC progress, increasing
DirAcc from $77.2\%$ before gating to $97.5\%$ among retained credits
(Figure~\ref{fig:credit-analysis}(b,e,f)).

\paragraph{Component ablations.}
\label{sec:ablations}
Removing historical evidence lowers average success from $98.3\%$ to
$94.0\%$, close to GRPO's $93.9\%$ (Table~\ref{tab:ablations}). Under our
settings, current-group evidence alone is insufficient for any credit to
pass the gate, so the update reduces to GRPO
(Appendix~\ref{app:pooled-endpoint-concentration}). Removing the gate also
lowers average success to $96.9\%$. Although it provides denser local
feedback, it admits uncertain credit directions into policy updates:
more credit does not necessarily provide better supervision.

\section{Conclusion}
\label{sec:conclusion}

We introduced PRICE, which derives action-chunk credit from terminal outcomes
through physical relations across trajectories. By pooling current and
historical evidence and applying confidence gating, PRICE refines
trajectory-level supervision without an auxiliary evaluator. Our analysis
connects oracle potential changes to the terminal-success objective and
\revision{provides a finite-sample directional bound for outcome-independent
evidence pools}. Experiments on LIBERO, RoboTwin~2.0, and real robots demonstrate
gains over outcome-based baselines, while controlled comparisons in simulation
show improved training efficiency. Independent continuation evaluation further
shows that retained credits align with local progress. Together, these results
show that physical correspondence across collected experience can turn sparse
terminal feedback into useful local supervision for embodied policy learning.
\clearpage

\bibliography{iclr2027_conference}
\bibliographystyle{iclr2027_conference}

\clearpage
\appendix
\numberwithin{equation}{section}
\makeatletter
\setlength{\@fptop}{0pt}
\setlength{\@fpsep}{12pt}
\setlength{\@fpbot}{0pt plus 1fil}
\makeatother

\begin{center}
    {\Large\bfseries Appendix}
\end{center}

\section{Endpoint Value Changes and the Terminal Objective}
\label{app:advantage-identity}

Under the assumptions of Theorem~\ref{thm:oracle-objective-alignment}, $H_t$
contains the task context, boundary observations, proprioception, time, and
previous chunks. It excludes other trajectories' outcomes and graph estimates.
The actor input $h_t$ is a fixed function of $H_t$. The terminal history
includes $Y$, so $V_{T+1}^\pi(H_{T+1})=Y$.

\paragraph{Proof of Theorem~\ref{thm:oracle-objective-alignment}.}
By the tower property,
\begin{align}
    \mathbb E_\pi[
      V_{t+1}^\pi(H_{t+1})\mid H_t,\mathbf a_t]
    &=
    \mathbb E_\pi[
      \mathbb E_\pi[Y\mid H_{t+1}]
      \mid H_t,\mathbf a_t] \nonumber\\
    &=\mathbb E_\pi[Y\mid H_t,\mathbf a_t].
    \label{eq:appendix-tower-identity}
\end{align}
Subtracting $V_t^\pi(H_t)$ proves
Equation~\ref{eq:oracle-progress-advantage}.

For the policy-gradient identity, fix $\theta_0$, set
$\pi=\pi_{\theta_0}$, and define
$S_t(\theta_0)=
\left.\nabla_\theta\log\pi_\theta(\mathbf a_t\mid h_t)
\right|_{\theta=\theta_0}$. Because the policy samples from
$\pi_{\theta_0}(\cdot\mid h_t)$,
$\mathbb E_{\pi_{\theta_0}}[S_t(\theta_0)\mid H_t]=0$, so
$V_t^{\pi_{\theta_0}}(H_t)$ is a valid baseline. The likelihood-ratio identity
and iterated expectation give
\begin{align}
    \left.\nabla_\theta J(\theta)\right|_{\theta=\theta_0}
    &=
    \mathbb E_{\pi_{\theta_0}}\!\left[
      \sum_{t=1}^{T}S_t(\theta_0)
      A_t^{\pi_{\theta_0}}(H_t,\mathbf a_t)
    \right] \nonumber\\
    &=
    \mathbb E_{\pi_{\theta_0}}\!\left[
      \sum_{t=1}^{T}S_t(\theta_0)
      \Delta_t^{\mathrm{ora}}
    \right].
    \label{eq:appendix-score-progress}
\end{align}
Stop-gradient implements this score-function estimator by differentiating only
the action probability.

\paragraph{Additive oracle credit.}
Let the expectation also include the on-policy rollout group used to compute
$\widehat A^{\mathrm{GRPO}}$ for the query trajectory. Define
$g_{\mathrm{base}}(\theta_0)=
\mathbb E_{\pi_{\theta_0}}[\sum_t S_t(\theta_0)\widehat A^{\mathrm{GRPO}}]$.
Linearity and Equation~\ref{eq:appendix-score-progress} give
\begin{equation}
    \begin{aligned}
    g_{\mathrm{aug}}^{\mathrm{ora}}(\theta_0)
      &=g_{\mathrm{base}}(\theta_0)
        +\lambda\mathbb E_{\pi_{\theta_0}}\!\left[
          \sum_t S_t(\theta_0)\Delta_t^{\mathrm{ora}}\right]\\
      &=g_{\mathrm{base}}(\theta_0)
        +\lambda\left.\nabla_\theta J(\theta)\right|_{\theta=\theta_0},
    \end{aligned}
    \label{eq:appendix-additive-oracle-credit}
\end{equation}
which proves Equation~\ref{eq:oracle-additive-alignment}.

\paragraph{Trajectory-wise identity.}
Consecutive endpoint values also cancel pathwise:
\begin{align}
    \sum_{t=1}^{T}\Delta_t^{\mathrm{ora}}
    &=\sum_{t=1}^{T}
      \left(V_{t+1}^{\pi_{\theta_0}}(H_{t+1})
      -V_t^{\pi_{\theta_0}}(H_t)\right) \nonumber\\
    &=Y-V_1^{\pi_{\theta_0}}(H_1).
    \label{eq:appendix-progress-telescoping}
\end{align}

\section{Pooled Endpoint Concentration}
\label{app:pooled-endpoint-concentration}

Fix a supported endpoint $x$ and $N_x>0$ distinct trajectory records, with
sample size and membership fixed independently of their terminal labels.
Reveal the records in collection order. Let $Y_{x,k}\in\{0,1\}$ be the outcome
of record $k$, and let $\mathcal F_{x,k}$ be an increasing pre-outcome
filtration containing its first-visit history and all earlier revealed
outcomes. Assume
$Y_{x,k}\mid\mathcal F_{x,k}\sim\operatorname{Bernoulli}(\mu_{x,k})$, where
\begin{equation}
    \mu_{x,k}
    =\mathbb E[Y_{x,k}\mid\mathcal F_{x,k}],
    \qquad
    \overline V_x=\frac{1}{N_x}\sum_{k=1}^{N_x}\mu_{x,k},
    \qquad
    \widehat V_x=\frac{1}{N_x}\sum_{k=1}^{N_x}Y_{x,k}.
    \label{eq:appendix-pooled-means}
\end{equation}
For any $\nu\in\mathbb R$, conditional Hoeffding's lemma gives
\begin{equation}
    \mathbb E\!\left[
      \exp\!\left(\nu(Y_{x,k}-\mu_{x,k})\right)
      \middle|\mathcal F_{x,k}
    \right]
    \leq \exp\!\left(\frac{\nu^2}{8}\right).
    \label{eq:appendix-hoeffding-lemma}
\end{equation}
Iterating this bound along the filtration and applying the Chernoff method to
both tails yields
\begin{equation}
    \Pr\!\left(
      \left|\widehat V_x-\overline V_x\right|\geq\epsilon
      \right)
    \leq 2\exp(-2N_x\epsilon^2).
    \label{eq:appendix-pooled-hoeffding}
\end{equation}
Substituting
$\epsilon=\sqrt{\log(4/\delta_{\mathrm{edge}})/(2N_x)}$ makes the right-hand
side $\delta_{\mathrm{edge}}/2$. Clipping the interval to $[0,1]$ preserves
coverage because both $\widehat V_x$ and $\overline V_x$ lie in $[0,1]$.
Therefore,
\begin{equation}
    \Pr\!\left(
      \overline V_x\notin[L_x,R_x]
    \right)
    \leq\frac{\delta_{\mathrm{edge}}}{2}.
    \label{eq:appendix-endpoint-coverage}
\end{equation}

For supported endpoints, the interval excludes zero exactly when
$|\widehat\Delta_{i,t}|>\epsilon_x+\epsilon_{x^+}$; clipping endpoint intervals
to $[0,1]$ preserves this criterion. With equal support
$N_x=N_{x^+}=N$ and observed contrast magnitude $d>0$, this requires
\begin{equation}
    N>\frac{2\log(4/\delta_{\mathrm{edge}})}{d^2}.
    \label{eq:gate-support-requirement}
\end{equation}
Under our parameter settings, $\delta_{\mathrm{edge}}=0.15$, so contrast
magnitudes $1.0$, $0.4$, and $0.3$ require at least 7, 42, and 73 supporting
trajectories per endpoint, respectively.
This motivates accumulating same-task evidence across rounds.
If evidence is restricted to a single group of $K=8$ trajectories, excluding
the query leaves at most seven peers. An endpoint with at most six peers
has radius at least $0.5231$, while seven peers give radius $0.4843$.
Thus, a gate could pass only if both endpoints have seven peers; these
endpoints then share the same peer set and have identical means. Such
within-group evidence alone yields no nonzero credit under the default gate.
Without historical outcome summaries, the augmented advantage therefore
reduces to the GRPO advantage.

\section{Pooled Finite-Sample Edge Direction}
\label{app:sampling-gate-proof}

{
\begin{corollary}[Direction bound for fixed evidence pools]
\label{thm:graph-contrast-resolution}
Fix a supported non-terminal edge. Suppose each endpoint pool has sample
size and membership fixed independently of its terminal labels and satisfies
the conditional Bernoulli assumptions in
Appendix~\ref{app:pooled-endpoint-concentration}. Then
$\overline\Delta_{i,t}\in\mathcal I_{i,t}$ with probability at least
$1-\delta_{\mathrm{edge}}$, and
\begin{equation}
    \Pr\!\left(
      C_{i,t}\neq0,\;
      \operatorname{sign}(C_{i,t})\neq
      \operatorname{sign}(\overline\Delta_{i,t})
    \right)\leq\delta_{\mathrm{edge}}.
    \label{eq:gated-credit-direction-error}
\end{equation}
\end{corollary}
\par}

\paragraph{Proof of \revision{Corollary}~\ref{thm:graph-contrast-resolution}.}
Apply Equation~\ref{eq:appendix-endpoint-coverage} to the source endpoint $x$
and destination endpoint $x^+$. A union bound gives
\begin{equation}
    \Pr\!\left(
      \overline V_x\in[L_x,R_x]
      \ \text{and}\
      \overline V_{x^+}\in[L_{x^+},R_{x^+}]
    \right)
    \geq1-\delta_{\mathrm{edge}}.
    \label{eq:appendix-joint-endpoint-coverage}
\end{equation}
No independence between the two endpoint estimates is needed. On this joint
event,
\begin{equation}
    \overline\Delta_{i,t}
    =\overline V_{x^+}-\overline V_x
    \in
    [L_{x^+}-R_x,\;R_{x^+}-L_x]
    =\mathcal I_{i,t}.
    \label{eq:appendix-edge-coverage}
\end{equation}
The empirical means lie in their own intervals by construction, so
$\widehat\Delta_{i,t}\in\mathcal I_{i,t}$ as well. If
$0\notin\mathcal I_{i,t}$, the interval lies strictly on one side of zero and
both contrasts have that sign. Since an accepted credit equals
$\widehat\Delta_{i,t}$, an incorrect nonzero credit implies failure of the
joint endpoint coverage event. Its probability is at most
$\delta_{\mathrm{edge}}$, proving
Equation~\ref{eq:gated-credit-direction-error} and
\revision{Corollary}~\ref{thm:graph-contrast-resolution}.

\revision{This is a marginal bound for a fixed edge under the stated
selection assumptions, not a bound conditional on emission or a simultaneous
training-wide guarantee. Historical KL screening and archive retention can
depend on terminal outcomes through policy updates, so the corollary does
not directly cover the adaptive training archive. Retained-credit direction
accuracy is evaluated under frozen-policy replay in
Section~\ref{sec:credit-alignment}.}
The endpoint union bound accommodates shared trajectories without requiring
independent estimates. P red or variance-adaptive concentration could exploit
this structure to obtain tighter intervals.

\subsection{From Pooled Evidence to Current-Policy Progress}
\label{app:credit-transfer}

Fix collection round $q$ with policy $\pi_q$. For each endpoint $x$, let
$(H_{x,k},t_{x,k})$ denote the first-visit history and boundary index of pooled
record $k$. Keeping these histories and their weights fixed, define the
current-policy reference potential and contrast as
\begin{equation}
    \begin{aligned}
    V_x^{q,\mathrm{pool}}
      &=\frac{1}{N_x}\sum_{k=1}^{N_x}
          V_{t_{x,k}}^{\pi_q}(H_{x,k}),\\
    \Delta_e^{q,\mathrm{pool}}
      &=V_{x^+}^{q,\mathrm{pool}}-V_x^{q,\mathrm{pool}}.
    \end{aligned}
    \label{eq:current-policy-pooled-reference}
\end{equation}
For query edge $e=(i,t)$, the oracle contrast is
$\Delta_e^{\mathrm{ora},q}
=V_{t+1}^{\pi_q}(H_{i,t+1})-V_t^{\pi_q}(H_{i,t})$.
The estimation error decomposes as
\begin{equation}
    \begin{aligned}
    \widehat\Delta_e-\Delta_e^{\mathrm{ora},q}
      &=\underbrace{\widehat\Delta_e-\overline\Delta_e}
          _{\text{sampling error}}\\
      &\quad+\underbrace{\overline\Delta_e-\Delta_e^{q,\mathrm{pool}}}
          _{\text{collection-policy mismatch}}\\
      &\quad+\underbrace{\Delta_e^{q,\mathrm{pool}}-\Delta_e^{\mathrm{ora},q}}
          _{\text{state-abstraction error}}.
    \end{aligned}
    \label{eq:credit-error-decomposition}
\end{equation}
\revision{Corollary}~\ref{thm:graph-contrast-resolution} controls the sampling term under
its evidence assumptions. The second term compares continuation policies at
the same recorded histories. The third measures the discrepancy between the
pooled histories and the query, including differences in physical state,
interaction history, and remaining horizon.

\paragraph{Direction transfer.}
Let
$B_e=|\overline\Delta_e-\Delta_e^{q,\mathrm{pool}}|
     +|\Delta_e^{q,\mathrm{pool}}-\Delta_e^{\mathrm{ora},q}|$.
\revision{The triangle inequality gives
$|\widehat\Delta_e-\Delta_e^{\mathrm{ora},q}|\leq E_e+B_e$.
For an accepted edge,}
\begin{equation}
    \revision{E_e+B_e<|\widehat\Delta_e|}
    \quad\Longrightarrow\quad
    \operatorname{sign}(C_e)
      =\operatorname{sign}(\Delta_e^{\mathrm{ora},q}).
    \label{eq:credit-direction-transfer}
\end{equation}
\revision{In particular, $E_e\leq r_e$ and
$B_e<|\widehat\Delta_e|-r_e$ suffice.}
The historical KL screen restricts the contributing collection rounds, and
physical matching determines which histories share evidence. Together with
the sampling interval, these choices govern the reliability of local credit.

\section{Implementation and Evaluation Details}
\label{app:implementation-evaluation}
\label{app:credit-analysis-metrics}

\subsection{PRICE Training Algorithm}
\label{app:training-lifecycle}

Visual descriptors use an independent frozen copy of the initial SFT encoder;
proprioception uses fixed checkpoint normalization statistics. Each round
computes credit from a frozen archive, updates the policy, and then commits
evidence from all raw rollout groups for the next round
(Algorithm~\ref{alg:price-training}).

\begin{algorithm}[H]
    \caption{PRICE training and history updates.}
    \label{alg:price-training}
    \small
    \algrenewcommand{\algorithmicrequire}{\textbf{Input:}}
    \begin{algorithmic}[1]
        \Require Initial policy $\pi_0$, fixed physical descriptor, and parameters
            $\eta$, $\delta_{\mathrm{edge}}$, $\lambda$, $d_{\max}$,
            $C_{\mathrm{hist}}$, $C_{\mathrm{node}}$
        \State Initialize task graphs $\mathcal G_0$, summaries $\mathcal S_0$,
            and cumulative policy KL $\xi_0=0$
        \For{each training round $q$}
            \State Freeze $\mathcal G_q$ and $\mathcal S_q$ as the query snapshot
            \State Collect grouped rollouts and terminal outcomes under fixed $\pi_q$
            \For{each rollout group $g$}
                \State Match frozen prototypes; cluster unmatched boundaries within $g$
                \State Normalize outcomes within $g$ to obtain $\widehat A_i^{\mathrm{GRPO}}$
                \For{each query trajectory $i\in g$}
                    \State Pool current-group leave-one-out counts with history satisfying
                        $D_{s,q}\leq d_{\max}$
                    \State Compute gated credit $C_{i,t}$ using
                        Equations~\ref{eq:bounded-history-node-value}--\ref{eq:price-credit}
                    \State $\widehat A_{i,t}^{\mathrm{PRICE}}\gets
                        \widehat A_i^{\mathrm{GRPO}}+\lambda C_{i,t}$
                \EndFor
            \EndFor
            \State Update $\pi_q$ to $\pi_{q+1}$ using
                $\widehat A^{\mathrm{PRICE}}$ in Equation~\ref{eq:grpo-surrogate}
            \State Update prototypes; cluster unmatched descriptors across groups of the same task
            \State Append one summary $(q,v,n_v,r_v,\xi_q)$ per visited node from all raw rollouts
            \State Apply the summary and node capacity limits $C_{\mathrm{hist}}$ and $C_{\mathrm{node}}$
            \State $\xi_{q+1}\gets\xi_q+\max\{0,\widehat{\operatorname{KL}}(\pi_q\|\pi_{q+1})\}$
            \State Publish $\mathcal G_{q+1}$ and $\mathcal S_{q+1}$ for queries in the next round
        \EndFor
    \end{algorithmic}
\end{algorithm}

\revision{Node matching does not explicitly use terminal labels}. Each trajectory contributes
once per node, using its first visit; temporary nodes use current-group evidence
only. The final chunk receives zero local credit, and advantages are not
renormalized after credit addition. Matching follows fixed trajectory--time
order within each group; archive updates follow task--group--trajectory--time
order.

Each summary stores its collection-round cumulative KL $\xi_{u_s}$, giving
$D_{s,q}=\xi_q-\xi_{u_s}$. Appendix~\ref{app:hyperparameters} specifies the
parameters and KL estimator. Nodes retain their most recent $C_{\mathrm{hist}}$
summaries; task graphs retain at most $C_{\mathrm{node}}$ nodes, evicting the
least recently matched. Archive changes are committed only after a successful
policy update and saved with actor checkpoints; failed updates discard staged
records.

\subsection{Training Settings}
\label{app:hyperparameters}

Table~\ref{tab:core-hyperparameters} summarizes simulation training settings,
PRICE credit parameters, and real-robot AWR weights. General training
settings are given in Section~\ref{sec:experimental-setup}, and real-robot
adaptation details are provided in Appendix~\ref{app:real-robot-details}.

\begin{table}[!t]
    \centering
    \small
    \caption{Core training hyperparameters. Optimization batch sizes count
    trajectories and are global across eight GPUs. Simulation and real-robot
    credit weights are reported separately.}
    \label{tab:core-hyperparameters}
    \setlength{\tabcolsep}{5pt}
    \begin{tabular}{@{}lc@{\hspace{1.5em}}lc@{}}
        \toprule
        Parameter & Setting & Parameter & Setting \\
        \midrule
        \multicolumn{2}{l}{\textit{Simulation training and GRPO}}
            & \multicolumn{2}{l}{\textit{PRICE in simulation}} \\
        Optimizer & AdamW & Credit weight $\lambda$ & $0.2$ \\
        Learning rate & $5\times10^{-6}$ & Matching threshold $\eta$ & $0.93$ \\
        Learning-rate schedule & Constant & \revision{Gate parameter} $\delta_{\mathrm{edge}}$ & $0.15$ \\
        Cases per raw rollout batch & 64 & Summaries per node $C_{\mathrm{hist}}$ & 4 \\
        Rollouts per case & 8 & Nodes per task $C_{\mathrm{node}}$ & 1,024 \\
        Trajectories per raw batch & 512 & Visual weight $w_{\mathrm{vis}}$ & $0.5$ \\
        Optimization mini-batch size & 128 & Proprioceptive weight $w_{\mathrm{prop}}$ & $0.5$ \\
        Optimization micro-batch size & 8 & Cumulative KL limit $d_{\max}$ & $0.2$ \\
        \cmidrule(l){3-4}
        GRPO epochs per update batch & 1
            & \multicolumn{2}{l}{\textit{Real-robot AWR adaptation}} \\
        Rollout temperature & $1.6$ & Credit weight $\lambda_{\mathrm{AWR}}$ & $0.2$ \\
        Gradient clipping norm & $1.0$ & Temperature $\tau_{\mathrm{AWR}}$ & $1.0$ \\
        Actions per chunk & 8 & Minimum weight $w_{\min}$ & $0.1$ \\
        Lower GRPO clip $\epsilon_{\mathrm{low}}$ & $0.20$ & Maximum weight $w_{\max}$ & $5.0$ \\
        Upper GRPO clip $\epsilon_{\mathrm{high}}$ & $0.28$ & Time-sampling shape $\alpha_{\mathrm{time}}$ & $1.5$ \\
        Reference KL coefficient $\beta$ & $0$ & Time-sampling shape $\beta_{\mathrm{time}}$ & $1.0$ \\
        & & Minimum flow time $u_{\min}$ & $0.001$ \\
        \bottomrule
    \end{tabular}
\end{table}

For historical-evidence eligibility, adjacent-policy KL is estimated from the
old-minus-post-update log probabilities on valid action tokens in the update
batch, averaged over microbatches and workers. Negative estimates are clipped
to zero before accumulation. The separate reference-policy KL penalty is
disabled ($\beta=0$). Real-robot AWR weights follow
Equation~\ref{eq:real-awr-weights}.

For ablations, PRICE w/o Historical Evidence sets $C_{\mathrm{hist}}=0$ while
retaining graph construction and gating; PRICE w/o Gate uses raw potential
differences for all supported non-terminal chunks; GRPO sets $\lambda=0$.

\subsection{Credit Evaluation Protocol}
\label{app:credit-evaluation-protocol}

\paragraph{Main credit audit.}
Figure~\ref{fig:credit-analysis} uses a frozen OpenVLA-OFT policy on
LIBERO-Spatial. The replay corpus contains five rollout rounds, 320 groups,
and 20,480 valid non-terminal action chunks. The audit uses $\eta=0.93$,
$\delta_{\mathrm{edge}}=0.15$, and up to four historical summaries per node,
covering 160 chunks in 16 windows selected before observing MC outcomes.
Table~\ref{tab:equal-coverage} and Figure~\ref{fig:credit-selection-distribution}
reuse this audit.

\paragraph{Controlled history and gate comparisons.}
The heatmaps share locked windows and endpoint node assignments.
Panel (d) uses terminal outcomes from current-group peers, excluding the
query trajectory and retaining each peer's first node visit. It computes
$\widehat V_x^{\mathrm{cur}}=r^{\mathrm{cur}}_{-i}(v)/n^{\mathrm{cur}}_{-i}(v)$
and displays the destination-minus-source difference before gating;
unsupported endpoints yield zero. Panel (e) uses the pooled potentials in
Equation~\ref{eq:bounded-history-node-value}; applying the gate in
Equation~\ref{eq:price-credit} yields (b). The training history ablation in
Table~\ref{tab:ablations} retains the gate.

\paragraph{Independent continuation evaluation.}
For an evaluated edge $e=(x,\mathbf a,x^+)$, we restore physical states
from the exact current-plus-historical evidence pools used in
Figure~\ref{fig:credit-analysis}(e) and continue the same frozen policy.
The main credit audit uses 12,695 continuations across 9,878 distinct
evidence states in 159 endpoint contexts. We first estimate each state's
continuation success rate, then average these rates with equal weight
within each endpoint pool, regardless of the number of continuations per
state. Their difference defines the independent MC estimate in panel (c):
\begin{equation}
    \widehat\Delta_e^{\mathrm{MC}}
    =\widehat V_{x^+}^{\mathrm{MC}}-\widehat V_x^{\mathrm{MC}}.
    \label{eq:mc-node-progress}
\end{equation}

\paragraph{Credit-quality metrics.}
Let $\mathcal S$ be an evaluated set of chunks with nonzero signed credits
$\widetilde C_e$ and MC differences $\widehat\Delta_e^{\mathrm{MC}}$.
Direction accuracy (DirAcc) is the percentage of sign matches on
$\mathcal S_{\neq}=\{e\in\mathcal S:\widehat\Delta_e^{\mathrm{MC}}\neq0\}$:
\begin{equation}
    \mathrm{DirAcc}(\mathcal S)
    =\frac{100}{|\mathcal S_{\neq}|}
      \sum_{e\in\mathcal S_{\neq}}
      \mathbb I\!\left[
      \operatorname{sign}(\widetilde C_e)=
      \operatorname{sign}(\widehat\Delta_e^{\mathrm{MC}})\right].
    \label{eq:credit-direction-accuracy}
\end{equation}
Spearman rank correlation (RankCorr) compares the ordering of signed
credits and MC differences:
\begin{equation}
    \mathrm{RankCorr}(\mathcal S)
    =\operatorname{Spearman}\!\left(
      (\widetilde C_e)_{e\in\mathcal S},
      (\widehat\Delta_e^{\mathrm{MC}})_{e\in\mathcal S}\right).
    \label{eq:credit-rank-correlation}
\end{equation}
Aligned MC gap (AlignedGap) averages the per-chunk aligned differences
$g_e=100\,\operatorname{sign}(\widetilde C_e)\widehat\Delta_e^{\mathrm{MC}}$,
in percentage points:
\begin{equation}
    \mathrm{AlignedGap}(\mathcal S)
    =\frac{1}{|\mathcal S|}\sum_{e\in\mathcal S}g_e.
    \label{eq:aligned-mc-effect}
\end{equation}
Positive $g_e$ indicates directional agreement and negative $g_e$
indicates disagreement. RankCorr and AlignedGap include MC ties.
Confidence intervals use rollout-group-cluster bootstrap resampling.

\paragraph{Evaluation subsets.}
Figure~\ref{fig:credit-analysis}(f) evaluates the 123 nonzero candidates
in (e), their 78 rejected candidates, and the 45 retained credits in (b).
All subsets compare pre-gate credit from (e) with the MC reference in (c).
For Table~\ref{tab:equal-coverage}, the correct/non-tied counts are 39/40
for confidence gating and 29/44 for endpoint-count selection.

\paragraph{Online archive audit.}
We audit three PRICE runs on LIBERO-Spatial at rounds 16, 32, and 63, with
$\delta_{\mathrm{edge}}=0.15$. At each stage, we uniformly sample 400 supported
non-terminal chunks per run, irrespective of gate acceptance. For retained
credits, we fix the selected evidence pools and estimate $\overline\Delta_e$
using 512 fresh continuations per endpoint pool from uniformly sampled
evidence histories under their collection policies. DirAcc, RankCorr, and
AlignedGap follow the definitions above, including the treatment of MC ties.
Results are reported in Table~\ref{tab:online-credit-alignment}.

\section{Real-Robot Experimental Details}
\label{app:real-robot-details}

\subsection{Robot Observations and Task Definitions}

Both tasks use the real-robot platform, task definitions, and deployment
configuration of HiFi-UMI \citep{ai2026hifi}. In Remote Insertion, the
robot grasps a remote control and inserts it into a target storage box. In
Produce Sorting, it places a designated fruit and vegetable into their
respective category-specific trays; an episode succeeds only if both
placements are completed. The policy receives camera images and a
20-dimensional proprioceptive state. The recorded executions
illustrated in Figure~\ref{fig:real-robot-composite}(b) contain four camera
streams, with upper and lower views from each wrist. The action prediction
horizons are 50 for Remote Insertion and 20 for Produce Sorting.

\subsection{Offline Adaptation Protocol}

We initialize every variant from an OpenPI-$\pi_{0.5}$ checkpoint post-trained
with 400 HiFi-UMI trajectories. Before adaptation, we run this initialization
policy on each task to collect 60 interaction trajectories. These fixed
task-specific datasets provide both the terminal outcomes and the physical
relations used for offline adaptation; no additional online rollouts are
introduced during optimization.

GRPO-AWR forms a trajectory-level advantage from task-normalized terminal
outcomes and assigns it to all action chunks in that trajectory. PRICE-AWR adds
the action-chunk relation credit to the same terminal advantage before mapping
the combined signal to a bounded AWR weight for the flow-matching loss. The two
methods share the same single-epoch schedule, random seed, global batch size of
4, and learning rate of $10^{-5}$.

\subsection{Offline Relational Credit and Weighted Flow Matching}
\label{app:real-robot-objective}

\paragraph{Fixed-buffer relational credit.}
Let $\mathcal D$ denote the training action chunks in a task-specific offline
buffer. We construct physical nodes once from frozen visual--proprioceptive
descriptors. Each distinct episode contributes its binary
terminal outcome once per visited node. For a query chunk $(i,t)$, episode
$i$ is excluded from both endpoint outcome estimates. The offline contrast
$c_{i,t}$ is the destination-minus-source potential difference for supported
non-terminal chunks, and zero for unsupported or final chunks. This
fixed-buffer adapter uses the raw contrasts without the online confidence
gate; graph assignments and credits remain fixed during adaptation.

\paragraph{Advantage construction.}
For task context $\kappa$, let $\bar Y_\kappa$ and $\sigma_\kappa$ be the mean
and population standard deviation of outcomes across distinct training
episodes. Let $\mathcal D_\kappa^{+}$ contain the training chunks whose
contrast is nonzero, and define their root-mean-square contrast by
\begin{equation}
    s_\kappa=
    \sqrt{\frac{1}{|\mathcal D_\kappa^{+}|}
      \sum_{(i,t)\in\mathcal D_\kappa^{+}}c_{i,t}^{2}}.
    \label{eq:real-credit-scale}
\end{equation}
We set $s_\kappa=1$ when this set is empty and use $\varepsilon=10^{-8}$ for
numerical stability. The offline advantages are
\begin{equation}
    \begin{aligned}
    A_i^{\mathrm{out}}
      &=\frac{Y_i-\bar Y_{\kappa(i)}}{\max(\sigma_{\kappa(i)},\varepsilon)},\\
    A_{i,t}^{\mathrm{AWR}}
      &=A_i^{\mathrm{out}}
        +\lambda_{\mathrm{AWR}}
          \frac{c_{i,t}}{\max(s_{\kappa(i)},\varepsilon)}.
    \end{aligned}
    \label{eq:real-awr-advantage}
\end{equation}
This scaling preserves zero credit. PRICE-AWR uses
$\lambda_{\mathrm{AWR}}=0.2$; GRPO-AWR uses only $A_i^{\mathrm{out}}$.

\paragraph{Bounded AWR weights.}
Index training chunks by $j$, and let $A_j$ denote the corresponding
method's advantage. We exponentiate, normalize by the training-buffer mean,
and clip the weights:
\begin{equation}
    \begin{aligned}
    r_j&=\exp(A_j/\tau_{\mathrm{AWR}}),
    &\bar r&=\frac{1}{|\mathcal D|}\sum_{j\in\mathcal D}r_j,\\
    w_j&=\operatorname{clip}(r_j/\bar r,w_{\min},w_{\max}),
    &\bar w&=\frac{1}{|\mathcal D|}\sum_{j\in\mathcal D}w_j.
    \end{aligned}
    \label{eq:real-awr-weights}
\end{equation}
We use $\tau_{\mathrm{AWR}}=1$, $w_{\min}=0.1$, and $w_{\max}=5$.
Each method computes its own $\bar r$ and $\bar w$, which remain fixed during
training. The loss uses $w_j/\bar w$ so that its scale is normalized by the
dataset mean after clipping.

\paragraph{Flow-matching objective.}
Both adaptation methods retain the OpenPI flow-matching time sampler, with
parameters reported in Table~\ref{tab:core-hyperparameters}.
For normalized action chunk $\mathbf a_j$, sample
$\boldsymbol\epsilon\sim\mathcal N(0,I)$ and
$b\sim\operatorname{Beta}(\alpha_{\mathrm{time}},\beta_{\mathrm{time}})$,
and set $u=u_{\min}+(1-u_{\min})b$.
The interpolated action is
$\mathbf x_u=(1-u)\mathbf a_j+u\boldsymbol\epsilon$, with target velocity
$\boldsymbol\epsilon-\mathbf a_j$.
Let $m_{j,r}$ mark valid action timesteps, and let $d=32$ be the padded model
action dimension. The per-chunk loss and weighted minibatch objective are
\begin{equation}
    \begin{aligned}
    \ell_j(\theta)
      &=\mathbb E_{u,\boldsymbol\epsilon}\!\left[
        \frac{\sum_r m_{j,r}
          \left\|v_\theta(h_j,\mathbf x_u,u)_r
            -(\boldsymbol\epsilon-\mathbf a_j)_r\right\|_2^2}
             {d\sum_r m_{j,r}}\right],\\
    \mathcal L_{\mathrm{AWR}}(\theta)
      &=\frac{1}{|\mathcal B|\bar w}
        \sum_{j\in\mathcal B}w_j\ell_j(\theta).
    \end{aligned}
    \label{eq:real-weighted-flow-objective}
\end{equation}
Here $\mathcal B$ contains chunks with at least one valid action timestep;
padded timesteps are masked out. All graph statistics, advantages, and weights
are fixed supervision signals. Both adaptation methods optimize this same
flow-matching objective, differing in whether relational credit enters the
AWR advantage.

\section{Additional Results and Analyses}
\label{app:additional-analyses}

\subsection{Run-to-Run Variation in Main Results}
  \label{app:main-results-std}

  Table~\ref{tab:main-results-std} reports mean success rates and
  sample standard deviations across three independent PRICE training
  runs. The Avg.\ standard deviation (SD) is computed from the average
  success rate within each run.

  \begin{table}[!t]
      \centering
      \scriptsize
      \caption{Main PRICE results (\%, mean $\pm$ SD) across three
      independent training runs.}
      \label{tab:main-results-std}
      \setlength{\tabcolsep}{3pt}

      \begin{tabular}{lccccc}
          \toprule
          \multicolumn{6}{l}{\textit{LIBERO}} \\
          Method & Spatial & Object & Goal & Long & Avg. \\
          \midrule
          PRICE
          & $99.2\pm{0.2}$
          & $99.6\pm{0.2}$
          & $99.4\pm{0.3}$
          & $97.6\pm{0.5}$
          & $99.0\pm{0.2}$ \\
          \bottomrule
      \end{tabular}

      \vspace{5pt}

      \begin{tabular}{lccccccc}
          \toprule
          \multicolumn{8}{l}{\textit{RoboTwin~2.0}} \\
          Method
          & \shortstack{Handover\\Block}
          & \shortstack{Lift\\Pot}
          & \shortstack{Move Can\\Pot}
          & \shortstack{Pick Dual\\Bottles}
          & \shortstack{Place Container\\Plate}
          & \shortstack{Place Empty\\Cup}
          & Avg. \\
          \midrule
          PRICE
          & $79.7\pm{2.3}$
          & $80.5\pm{1.6}$
          & $90.6\pm{1.6}$
          & $94.5\pm{1.4}$
          & $97.7\pm{0.8}$
          & $96.9\pm{0.8}$
          & $90.0\pm{0.8}$ \\
          \bottomrule
      \end{tabular}
  \end{table}
  
\subsection{Training Efficiency}
\label{app:training-gpu-hours}

GPU hours (GPUh) are computed as the number of GPUs multiplied by elapsed
training time in hours. Table~\ref{tab:training-gpu-hours} reports cumulative
cost at the 90\% and 95\% rollout-success milestones in
Figure~\ref{fig:training-efficiency}(a--b).
At the 90\% milestone, PRICE uses 272.8 GPU hours on Goal and 221.2 on
Spatial, reducing cost by $38.3\%$ and $32.8\%$, respectively, relative to
SimpleVLA-RL. At 95\%, the reductions are $17.3\%$ and $23.8\%$.
These results show that PRICE's gains in training-step efficiency also
translate into lower cumulative GPU-hour costs.

\begin{table}[!t]
    \centering
    \small
    \caption{Cumulative GPU hours at the marked milestones in
    Figure~\ref{fig:training-efficiency}(a--b). Reductions are relative to
    SimpleVLA-RL.}
    \label{tab:training-gpu-hours}
    \setlength{\tabcolsep}{6pt}
    \begin{tabular}{lcccc}
        \toprule
        Suite & Success (\%) & SimpleVLA-RL & PRICE & Reduction (\%) \\
        \midrule
        Goal & 90 & 441.9 & \textbf{272.8} & 38.3 \\
             & 95 & 625.2 & \textbf{517.2} & 17.3 \\
        \midrule
        Spatial & 90 & 329.0 & \textbf{221.2} & 32.8 \\
                & 95 & 714.6 & \textbf{544.3} & 23.8 \\
        \bottomrule
    \end{tabular}
\end{table}

\subsection{Mechanism Analysis}
\label{app:mechanism-analysis}

\paragraph{Credit alignment during online training.}
Retained credits achieve mean DirAcc above $94\%$, positive rank correlations,
and positive aligned MC gaps at all three training stages
(Table~\ref{tab:online-credit-alignment}). These results support agreement
between online PRICE credit and the independent collection-policy MC
reference as the policy and archive evolve. The audit protocol is detailed
in Appendix~\ref{app:credit-evaluation-protocol}.

\begin{table}[htbp]
    \centering
    \small
    \caption{Online credit alignment on LIBERO-Spatial. Of 1,200 audited
    chunks per stage across three PRICE runs, 24, 48, and 36 receive nonzero
    credit at rounds 16, 32, and 63, respectively.}
    \label{tab:online-credit-alignment}
    \setlength{\tabcolsep}{6pt}
    \begin{tabular}{lccc}
        \toprule
        Training round & DirAcc (\%) $\uparrow$ & RankCorr $\uparrow$
            & AlignedGap (pp) $\uparrow$ \\
        \midrule
        16 & 95.8 & 0.67 & +24.6 \\
        32 & 95.8 & 0.71 & +27.8 \\
        63 & 94.3 & 0.68 & +26.2 \\
        \bottomrule
    \end{tabular}
\end{table}

\paragraph{Physical context of credited chunks.}
We analyze the physical context of action chunks assigned nonzero credit by
PRICE in LIBERO-Spatial. Each chunk is assigned to one of four mutually
exclusive categories, in order: (i) gripper--target contact, detected at a
chunk boundary, or target-object displacement greater than 1 cm; (ii) an
arm-induced reduction of more than 1 cm in gripper-to-object surface distance,
provided category (i) does not apply; (iii) a source or destination
configuration within 5 cm of the target-object surface, provided neither of
the first two categories applies; and (iv) other. The first three categories
constitute interaction-related intervals. Of the 777 credited chunks, 572
(73.6\%) fall within these intervals, accounting for 72.2\% of the total
absolute credit magnitude. Contact or manipulation, approach, and proximity
alone account for 33.7\%, 19.4\%, and 20.5\% of all credited chunks,
respectively. Thus, most credited chunks occur during approach to, contact
with, or positioning near the target object, as well as object movement.

\paragraph{Credit selection at equal coverage.}
We examine whether confidence gating selects more reliable credits than
random subsampling or selection based on evidence quantity. All three
selectors retain 45 of the same 123 nonzero candidates in
Figure~\ref{fig:credit-analysis}. Endpoint-count selection keeps candidates
with the largest minimum endpoint support, $\min(N_x,N_{x^+})$, while
random selection samples uniformly without replacement. Confidence gating
achieves $97.5\%$ DirAcc and $+28.5$ pp AlignedGap, outperforming both
controls (Table~\ref{tab:equal-coverage}). These results support assessing
potential changes relative to their uncertainty when selecting local credit.

\begin{table}[!t]
    \centering
    \small
    \caption{Credit selection at equal coverage on LIBERO-Spatial. Each
    selector retains 45 of the same 123 candidates. MC ties are excluded from
    DirAcc; the random row reports means over 100,000 subsets.}
    \label{tab:equal-coverage}
    \setlength{\tabcolsep}{6pt}
    \begin{tabular}{lcc}
        \toprule
        Selector & DirAcc (\%) $\uparrow$
            & AlignedGap (pp) $\uparrow$ \\
        \midrule
        Confidence gate & \textbf{97.5} & \textbf{+28.5} \\
        Endpoint-count selection & 65.9 & +4.6 \\
        Random selection & 77.2 & +12.8 \\
        \bottomrule
    \end{tabular}
\end{table}

\begin{figure}[!t]
    \centering
    \includegraphics[width=\linewidth]{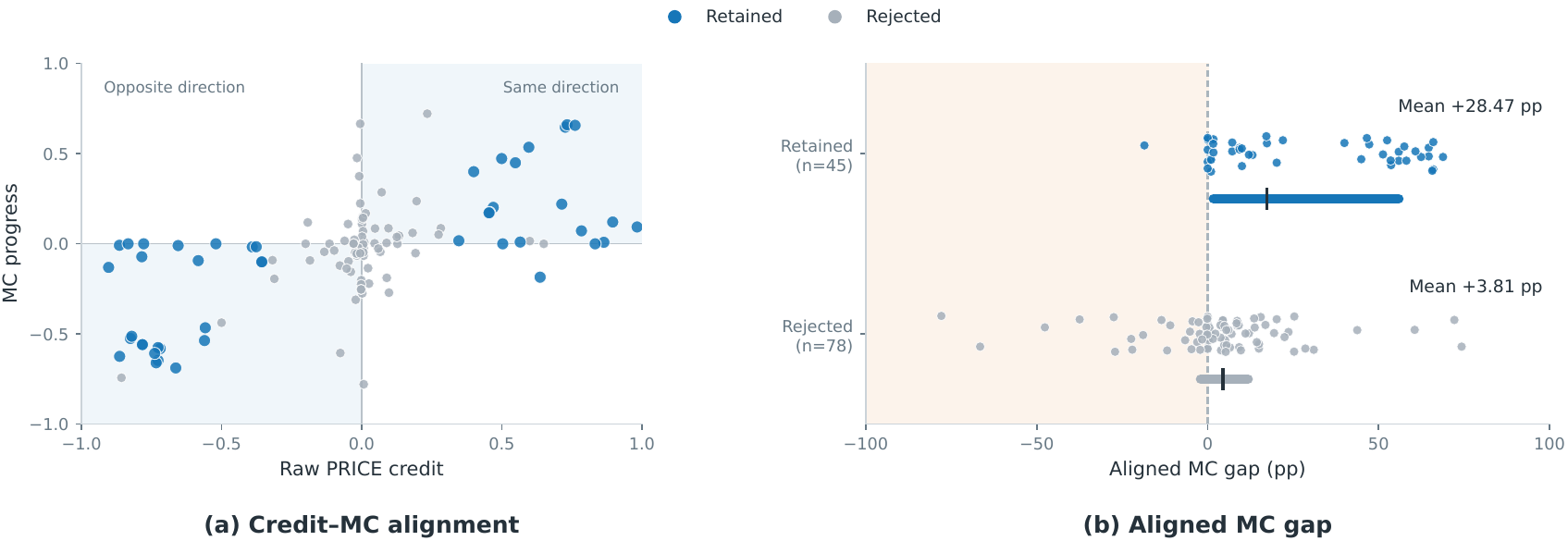}
    \caption{Credit selection and local progress on LIBERO-Spatial.
    (a) Raw PRICE credit versus MC progress for the same 123 candidates as
    Figure~\ref{fig:credit-analysis};
    blue and gray denote the 45 retained and 78 rejected candidates,
    respectively. Shaded quadrants indicate agreement in sign.
    (b) Per-chunk aligned MC gaps $g_e$ (pp), whose means give AlignedGap.
    Vertical jitter separates points;
    thick segments show interquartile ranges, dark ticks mark medians, and
    annotations report means.}
    \label{fig:credit-selection-distribution}
\end{figure}

\paragraph{Alignment of retained and rejected credits.}
We examine whether gating retains credits that better reflect local progress.
Figure~\ref{fig:credit-selection-distribution}(a) compares pre-gate credit
with independent MC estimates for the same candidates as
Figure~\ref{fig:credit-analysis}. Retained credits exhibit a higher rate of
directional agreement with MC progress than rejected candidates.
Figure~\ref{fig:credit-selection-distribution}(b) additionally accounts for
the magnitude of MC changes by aligning them with each candidate's credit
direction. Retained credits achieve an AlignedGap of $+28.47$ pp, compared
with $+3.81$ pp for rejected candidates. Their aligned-gap distribution has
an interquartile range above zero, whereas the rejected subset's spans zero.
These results indicate that gating selects credits with more consistent
directional agreement and larger average MC changes in the credited direction.

\subsection{Descriptor Robustness}
\label{app:representation-robustness}

\paragraph{PRICE remains reliable across descriptor choices.}
In an independent experiment on 20,480 action chunks, we rebuild the graph
using proprioceptive, fused visual--proprioceptive, and visual descriptors
with the remaining settings fixed. All three yield DirAcc above $87\%$ and positive AlignedGap
(Table~\ref{tab:representation-robustness}). Fusion achieves the highest
credit-quality scores: $93.9\%$ DirAcc, $0.758$ RankCorr, and $+33.7$ pp
AlignedGap. Vision provides the highest coverage at $4.2\%$, compared
with $3.8\%$ for Fusion and $3.6\%$ for Proprio.

\begin{table}[!t]
    \centering
    \small
    \caption{Independent descriptor comparison. Coverage is the
    percentage of the 20,480 corpus chunks receiving nonzero credit;
    quality metrics use descriptor-specific MC samples and the definitions
    in Appendix~\ref{app:credit-evaluation-protocol}. Brackets give $95\%$
    rollout-group-cluster bootstrap intervals.}
    \label{tab:representation-robustness}
    \setlength{\tabcolsep}{3pt}
    \begin{tabular}{lcccc}
        \toprule
        Descriptor & Coverage & DirAcc $\uparrow$ & RankCorr $\uparrow$
            & AlignedGap $\uparrow$ \\
                   & (\%) & (\%) & & (pp) \\
        \midrule
        Proprio  & 3.6 & 88.2 & 0.650 & +23.4 \\
                 &       & $[75.7,97.4]$ & & $[13.6,33.0]$ \\
        Vision   & 4.2 & 87.2 & 0.595 & +29.0 \\
                 &       & $[71.4,100.0]$ & & $[19.6,38.5]$ \\
        Fusion   & 3.8 & \textbf{93.9} & \textbf{0.758}
            & \textbf{+33.7} \\
                 &       & $[83.9,100.0]$ & & $[23.7,43.5]$ \\
        \bottomrule
    \end{tabular}
\end{table}

\newpage
\subsection{Hyperparameter Sensitivity}
\label{app:hyperparameter-sensitivity}

\paragraph{Sensitivity to the credit weight.}
After 63 training steps on LIBERO-Spatial, $\lambda=0.2$ and $0.4$
achieve similar success ($97.0\%$ and $96.9\%$, respectively;
Table~\ref{tab:credit-weight-sensitivity}). However, $\lambda=0.2$
reaches $90\%$ success in 22 steps versus 29 for $\lambda=0.4$,
supporting our default choice for its faster learning at comparable
final success.

\begin{table}[!t]
    \centering
    \small
    \caption{Credit-weight sensitivity on LIBERO-Spatial.
    Task success is evaluated after 63 training steps; bold marks the default weight.}
    \label{tab:credit-weight-sensitivity}
    \setlength{\tabcolsep}{10pt}
    \begin{tabular}{lcccc}
        \toprule
        Credit weight $\lambda$ & 0.1 & \textbf{0.2} & 0.3 & 0.4 \\
        \midrule
        Success (\%) & 95.8 & \textbf{97.0} & 96.5 & 96.9 \\
        \bottomrule
    \end{tabular}
\end{table}

\paragraph{Credit selection across gate settings.}
  Table~\ref{tab:appendix-gate} varies $\delta_{\mathrm{edge}}$
  on 160 locked chunks while holding node assignments and evidence
  pools fixed. The stricter settings $0.05$ and $0.10$ retain 25
  and 30 credits, respectively, with $100\%$ DirAcc on non-tied
  MC comparisons. At the default setting of $0.15$, the gate
  retains 45 credits with $97.5\%$ DirAcc and $+28.5$ pp
  AlignedGap. Relaxing the gate to $0.30$ retains 46 credits,
  while DirAcc decreases to $95.1\%$ and AlignedGap to
  $+27.3$ pp.

\begin{table}[!t]
    \centering
    \small
    \caption{Gate-parameter sensitivity on 160 locked chunks.
    Retained counts nonzero credits; bold marks the \revision{default setting}.
    Metric definitions are given in Appendix~\ref{app:credit-evaluation-protocol}.}
    \label{tab:appendix-gate}
    \setlength{\tabcolsep}{5pt}
    \begin{tabular}{cccc}
        \toprule
        $\delta_{\mathrm{edge}}$ & Retained
            & DirAcc (\%) $\uparrow$ & AlignedGap (pp) $\uparrow$ \\
        \midrule
        0.05 & 25 & 100.0 & +30.8 \\
        0.10 & 30 & 100.0 & +29.3 \\
        \textbf{0.15} & 45 & 97.5 & +28.5 \\
        0.30 & 46 & 95.1 & +27.3 \\
        \bottomrule
    \end{tabular}
\end{table}

\paragraph{Choice of the matching threshold.}
Table~\ref{tab:appendix-graph} compares credit coverage and archive size
across matching thresholds, with other parameters fixed to the defaults in
Table~\ref{tab:core-hyperparameters}. We use $\eta=0.93$ to obtain the
highest observed coverage of gated credit with a relatively small archive:
777 credited chunks and 484 nodes. Higher tested thresholds increase
archive size while reducing coverage.

\begin{table}[!t]
    \centering
    \small
    \caption{Credit coverage and archive size under frozen-policy replay of
    the same 20,480 chunks. Credited chunks counts nonzero gated credits;
    Nodes and Summaries are final archive totals across tasks.
    Evictions counts cumulative node removals due to the 1,024-node limit
    per task.}
    \label{tab:appendix-graph}
    \begin{tabular}{lrrrr}
        \toprule
        $\eta$ & Credited chunks & Nodes & Summaries & Evictions \\
        \midrule
        0.90 & 628 & 287 & 873 & 0 \\
        0.93 & \textbf{777} & 484 & 1,405 & 0 \\
        0.96 & 604 & 1,234 & 3,064 & 0 \\
        0.99 & 302 & 7,458 & 12,325 & 4,678 \\
        \bottomrule
    \end{tabular}
\end{table}

\section{Extension to World Action Models}
\label{app:wam-extension}

To examine applicability beyond VLA policies, we study PRICE on
Fast-WAM~\citep{yuan2026fast}, a representative world action model that
combines video modeling with action generation. We compare GRPO and PRICE
on LIBERO-10 to assess the benefit of relational credit with a WAM backbone.

\paragraph{Initialization.}
We follow the official Fast-WAM SFT pipeline and select the
\texttt{step\_006000.pt} checkpoint, which achieves $53.4\%$ success on
LIBERO-10, as the base model for both GRPO and PRICE.

\paragraph{Policy adaptation.}
Fast-WAM generates continuous action means through flow matching. For each
rollout chunk, we cache the features $z_{i,t}$ and action latent $x_{i,t}$
entering the final flow step, together with its integration increment
$\Delta s$. Writing $\xi_{i,t}=(z_{i,t},x_{i,t},\Delta s)$, the update uses
\begin{equation}
    \begin{aligned}
    \mu_\theta(\xi_{i,t})
        &=\mathcal P\!\left(x_{i,t}+\Delta s\,f_\theta(z_{i,t})\right),\\
    q_\theta(\mathbf a\mid\xi_{i,t})
        &=\mathcal N\!\left(\mu_\theta(\xi_{i,t}),
            \operatorname{diag}(\sigma_\theta^2)\right),
    \end{aligned}
    \label{eq:wam-gaussian-adaptation}
\end{equation}
where $f_\theta$ is the action output head and $\mathcal P$ applies action
denormalization, environment-specific postprocessing, and selection of the
executed action horizon. Rollouts sample from this diagonal Gaussian;
updates hold $\xi_{i,t}$ fixed and optimize the output head and log standard
deviations, keeping the remaining backbone frozen.
For action coordinate $k$ within a chunk, the likelihood ratio is
\begin{equation}
    \rho_{i,t,k}(\theta)
    =\frac{\mathcal N\!\left(a_{i,t,k};
        \mu_{\theta,k}(\xi_{i,t}),\sigma_{\theta,k}^{2}\right)}
        {\mathcal N\!\left(a_{i,t,k};
        \mu_{\theta_{\mathrm{old}},k}(\xi_{i,t}),
        \sigma_{\theta_{\mathrm{old}},k}^{2}\right)}.
    \label{eq:wam-coordinate-ratio}
\end{equation}
We use these ratios in the clipped surrogate of
Equation~\ref{eq:grpo-surrogate}, averaging over valid action coordinates
and sharing each chunk's advantage across its coordinates.

\paragraph{Controlled comparison.}
Both methods share the initial checkpoint, rollout grouping, optimizer
settings, training budget, and evaluation protocol. GRPO uses the
trajectory-level advantage, while PRICE applies the augmentation in
Equation~\ref{eq:price-augmented-advantage}. PRICE uses visual--proprioceptive
boundary descriptors, cross-trajectory outcome pooling, and confidence-gated
credit with the settings in Table~\ref{tab:core-hyperparameters}; credit
addition is not followed by advantage renormalization.
Table~\ref{tab:fastwam-libero10} reports aggregate success rates and includes
the official Fast-WAM full-SFT result of $95.2\%$ for reference.
PRICE achieves $97.2\%$ success, compared with $92.8\%$ for GRPO, an
improvement of $4.4$ percentage points. This result supports
the applicability of relational credit beyond VLA policies to a WAM-based
control policy.

\begin{table}[!t]
    \centering
    \small
    \caption{Success rates (\%) on LIBERO-10 with Fast-WAM, evaluated over
    50 episodes per task. The official full-SFT result is included for reference.}
    \label{tab:fastwam-libero10}
    \setlength{\tabcolsep}{10pt}
    \begin{tabular}{lc}
        \toprule
        Method & LIBERO-10 success (\%) \\
        \midrule
        Fast-WAM (official full SFT) & 95.2 \\
        \midrule
        SFT initialization (step 6,000) & 53.4 \\
        + GRPO & 92.8 \\
        + PRICE & \textbf{97.2} \\
        \bottomrule
    \end{tabular}
\end{table}

\section{Additional Visualizations}
\label{app:additional-visualizations}

We show chronological execution sequences from simulation and the real robot.

\subsection{Simulated Manipulation}
\label{app:simulation-visualizations}

\begin{figure}[!t]
    \centering
    \includegraphics[width=\linewidth]{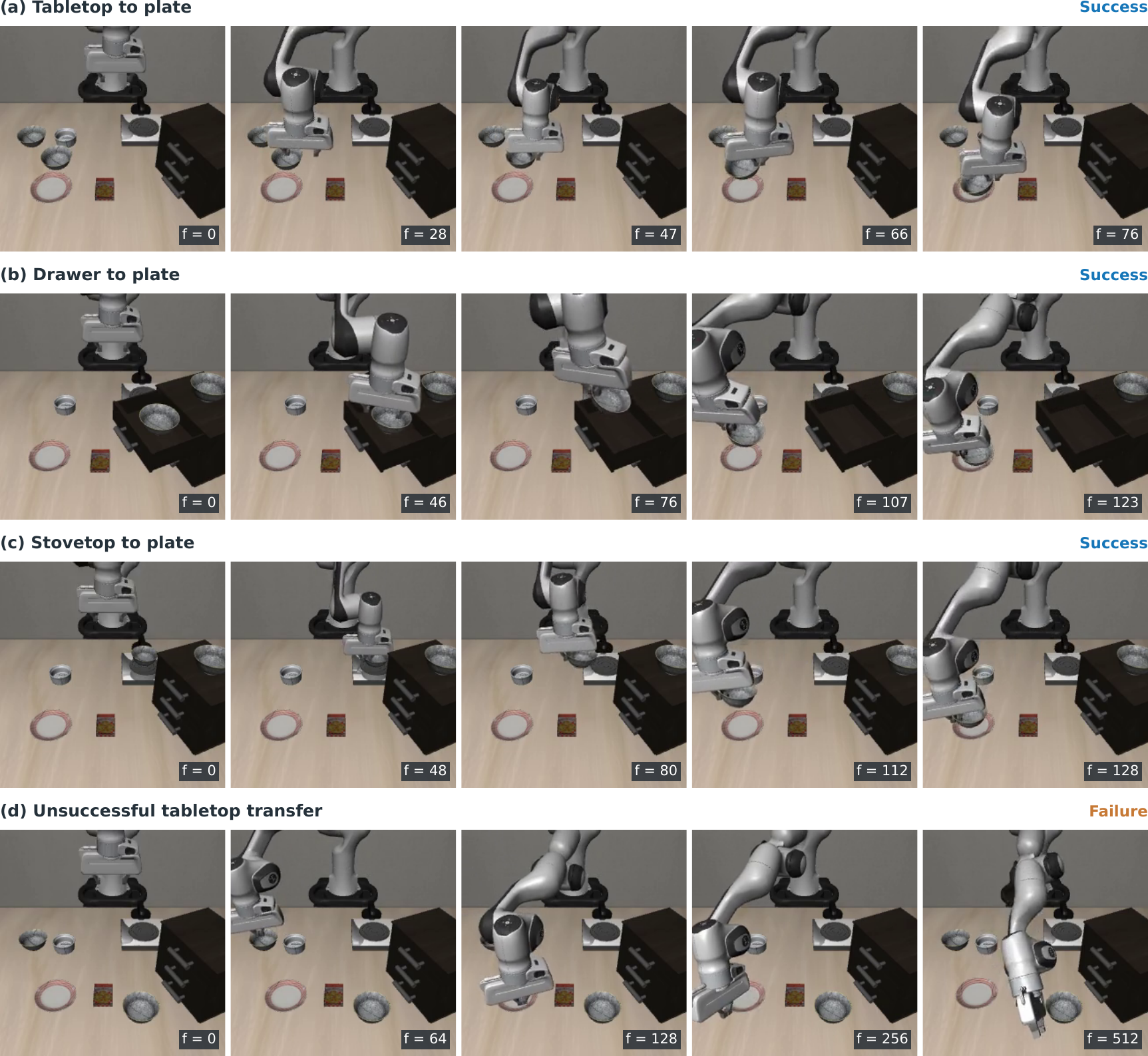}
    \caption{Additional simulated execution examples on LIBERO-Spatial.
    Each row shows five chronological frames from one recorded rollout:
    successful bowl transfers from (a) the tabletop, (b) a drawer, and
    (c) the stovetop, followed by (d) an unsuccessful tabletop transfer.
    Labels $f$ are zero-based video frame indices.}
    \label{fig:additional-simulation-executions}
\end{figure}

Figure~\ref{fig:additional-simulation-executions} shows four recorded
OpenVLA-OFT executions on LIBERO-Spatial. Different bowl starting locations
require different extraction and transport motions before placement on the
plate. The unsuccessful attempt reaches the plate region but does not
complete the task.

\clearpage
\subsection{Real-Robot Manipulation}
\label{app:real-robot-visualizations}

\begin{figure}[!t]
    \centering
    \includegraphics[width=\linewidth]{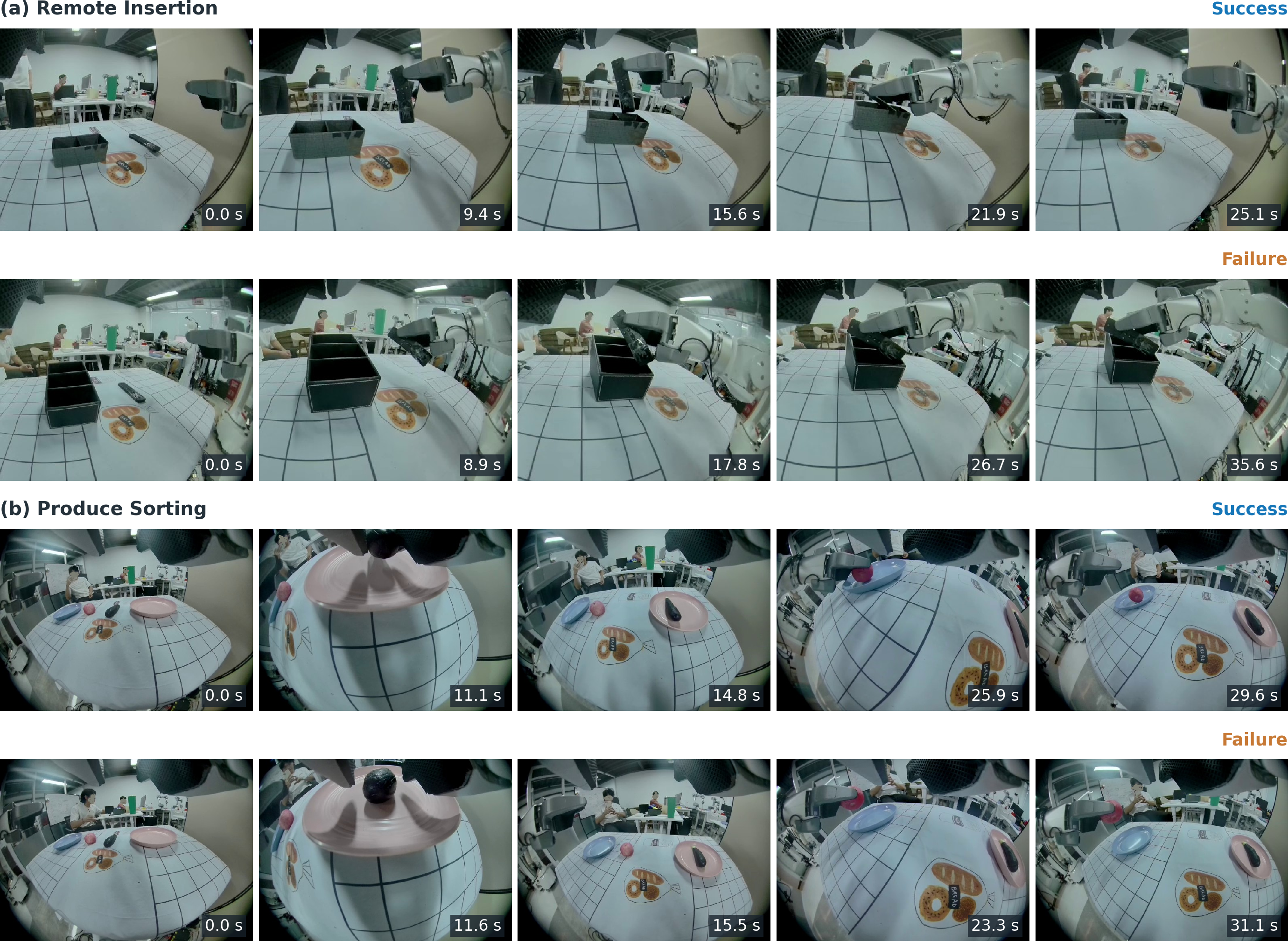}
    \caption{Additional real-robot execution examples.
    (a) Remote Insertion and (b) Produce Sorting each show one successful
    episode (upper row) and one unsuccessful episode (lower row), with
    five chronological frames per episode. Timestamps are relative to
    episode start. Remote Insertion uses the lower left-wrist camera;
    Produce Sorting uses the lower right-wrist camera to show both trays.
    Outcomes are operator-provided episode labels.}
    \label{fig:additional-real-robot-executions}
\end{figure}

Figure~\ref{fig:additional-real-robot-executions} presents four additional
episodes, distinct from those in Figure~\ref{fig:real-robot-composite}(b).
The unsuccessful insertion attempt keeps the remote near the box, while
the unsuccessful sorting attempt completes the vegetable placement but
leaves the fruit held near its tray. The successful sorting sequence shows
both placements, extending the fruit-only view in the main figure.

\clearpage
\section{Limitations}
\label{app:limitations}

PRICE trades off evidence reuse against collection-policy mismatch and
state-abstraction error (Appendix~\ref{app:credit-transfer}). Credit quality
depends on how well the descriptor preserves distinctions relevant to success
and how representative historical outcomes remain under the current policy.
Short observation histories could improve physical correspondence, while
more targeted history selection could reduce policy mismatch. Local credit also
requires repeated visits to corresponding states, limiting its coverage
during early training or in tasks with little overlap across trajectories.
Improved exploration and evidence reuse could expand this coverage.
Finally, training becomes less efficient as success rates approach
saturation. Further gains beyond 90--95\% require additional interaction,
while increasingly uniform terminal outcomes reduce the diversity of new
evidence for credit assignment. Sampling difficult tasks or initial
conditions more frequently could concentrate the interaction budget on
remaining failures and improve learning efficiency in this regime.

\end{document}